\documentclass{article}

\usepackage[preprint]{neurips_2024}
\usepackage{algorithm}
\usepackage{algpseudocode}
\usepackage{graphicx}
\usepackage{subcaption}
\usepackage{wrapfig}
\usepackage{tikz}
\usepackage{float} 

\usepackage{color}

\usepackage[utf8]{inputenc} 
\usepackage[T1]{fontenc}    
\usepackage{url}            
\usepackage{booktabs}       
\usepackage{amsfonts}       
\usepackage{nicefrac}       
\usepackage{microtype}      
\usepackage{xcolor}         
\usepackage[english]{babel}
\usepackage{amsmath}

\usepackage{hyperref}       

\newcommand{\br}[1]{\left\{#1\right\}}
\newcommand{\len}{\operatorname{len}}
\newcommand{\pr}[1]{\left(#1\right)}
\newcommand{\clip}{\operatorname{clip}}
\newcommand{\PPO}{\operatorname{PPO}}
\newcommand{\RRR}{\operatorname{RRR}}
\newcommand{\fit}{\operatorname{fit}}

\title{Rewarded Region Replay (R3) for Policy Learning with Discrete Action Space}

\author{%
    Bangzheng Li$^{*}$ \\
    MIT \\
    \texttt{liben@mit.edu} \\
    \And
    Ningshan Ma \\
    MIT\\
    \texttt{ningshan@mit.edu} \\
    \AND
    Zifan Wang \\
    MIT \\
    \texttt{atticusw@mit.edu} \\
}

\begin{document}

\maketitle

\begin{abstract}
We introduce a new on-policy algorithm called Rewarded Region Replay (R3), which significantly improves on PPO in solving environments with discrete action spaces. R3 improves sample efficiency by using a replay buffer which contains past successful trajectories with reward above a certain threshold, which are used to update a PPO agent with importance sampling. Crucially, we discard the importance sampling factors which are above a certain ratio to reduce variance and stabilize training. We found that R3 significantly outperforms PPO in Minigrid environments with sparse rewards and discrete action space, such as DoorKeyEnv and CrossingEnv, and moreover we found that the improvement margin of our method versus baseline PPO increases with the complexity of the environment. We also benchmarked the performance of R3 against DDQN (Double Deep Q-Network), which is a standard baseline in off-policy methods for discrete actions, and found that R3 also outperforms DDQN agent in DoorKeyEnv. Lastly, we adapt the idea of R3 to dense reward setting to obtain the Dense R3 algorithm (or DR3) and benchmarked it against PPO on Cartpole-V1 environment. We found that DR3 outperforms PPO significantly on this dense reward environment. Our code can be found at \url{https://github.com/chry-santhemum/R3}.

\end{abstract}

\section{Introduction}
Navigating environments with sparse rewards is a significant challenge in reinforcement learning, particularly for traditional algorithms like Proximal Policy Optimization (PPO) and DDQN (Double Deep Q network). Receiving rewards in such environments is an infrequent event, making the learning task much harder. 

Current on-policy and off-policy methods each have their pros and cons. For example, on-policy algorithms like PPO are usually more stable and faster than off-policy algorithms, and off-policy algorithms like DDQN are more sample-efficient than on-policy algorithms due to the existence of a replay buffer. As a result, in sparse reward settings, off-policy algorithms usually performs better than on-policy algorithm due to the increase sample efficiency.

We propose and implement an algorithm which we call Rewarded Region Replay (or R3) which combines the pros of on-policy and off-policy algorithms: it is stable and runs fast, while at the same time is sample efficient due to the existence of a replay buffer, and thus better at difficult tasks with sparse rewards.

One starting point of R3 is to mimic how humans would develop strategies in a sparse rewards environment. When we are performing tasks in which the result we get is either success or failure, we may experience a long period of failure before reaching a state of success. However, when we successfully solve the task, we would reflect on what has made us win this one time and apply it to future trials. In other words, we add the successful data into a replay buffer and think back about what made this data successful.

R3 mimics this process by creating a replay buffer for an on-policy algorithm like PPO. This replay buffer only stores those successful experiences that yield positive reward. However, there is a well-known problem in using a replay buffer for on-policy algorithms: distribution shift. As the policy changes, the current policy distribution may no longer agree with the trajectories in the replay buffer.

In order to solve this problem, the natural solution is to use importance sampling. However, naive importance sampling poses a critical flaw: its variance is too high. In importance sampling, we are multiplying each term by the ratio of the probability produced by the new policy and the probability produced by the old policy. If this ratio is too high, then in the gradient descent step, the parameters would change too rapidly and the policy may not converge. We deal with this problem by discarding the terms where the ratios that are too high in the gradient descent. The adjusted importance sampling combined with the replay buffer is the base of our R3 algorithm.

\section{Related Work}

\subsection{Proximal policy optimization (PPO)}

Since the introduction of proximal policy optimization (PPO) in 2017 by Schulman et al. \cite{schulman2017proximal}, it has found major successes over a wide range of tasks from playing Dota 2 \cite{openai2019dota} to aligning language models \cite{ouyang2022training}. However, even on simple environments such as Minigrid DoorKey, there are already several limitations. Due to the sparsity of reward signals in such environments, significant interactions are rare, and empirically it could take a relatively long time before the agent starts getting enough rewards to learn from them. Furthermore, hyperparameters such as the clip ratio, learning rate, and entropy coefficient are quite sensitive to the learning landscape, and often require scheduling or annealing.

\subsection{Double deep Q-network (DDQN)}
Double deep Q-network (DDQN), introduced in \cite{vanhasselt2015deep}, 
addresses the overestimation bias of Q-values found in standard deep Q-networks (DQN) by decoupling the selection and evaluation of the action-value function. To make DDQN more sample-efficient, Schaul et al. introduced a method called prioritized experience replay \cite{schaul2016prioritized}, in which more informative transitions are sampled more frequently. In another direction, Andrychowicz et al. had introduced hindsight experience replay (HER) \cite{andrychowicz2018hindsight}, which automatically sets intermediate goals in failure trajectories and replays them, so that the agent also learns from experiences without external rewards. HER can be combined with any off-policy method (such as DDQN), and is currently considered as one of the best algorithms for sparse reward tasks.

\subsection{Experience replay}

Experience replay is a familiar idea \cite{experienceReplay}, but it has inspired many innovations in modern machine learning. In tasks with sparse rewards, the role of a replay buffer is more important, because it allows for repeated sampling of rarely occurring but useful transitions. Previously discussed limitations of PPO, as well as off-policy methods such as HER, inspired us to use experience replay in on-policy settings. Thus, we propose to add to PPO a replay buffer containing interactions that successfully found rewards, encouraging the agent to learn more from them using importance sampling. In the next section, we expand on this idea and mention several crucial details in implementation that improve on the performance.

\section{Methods}

\subsection{Environment}

The environments on which we performed our evaluation of R3 are the Crossing and DoorKey minigrid environments and the CartPole v1 gym environment.
\begin{itemize}
    \item The Crossing environment consists of a grid, an agent, and a goal, and the objective is to navigate to the goal location. There is also one stream of Lava running either horizontally or vertically, leaving exactly one block that is open to crossing; the episode ends if the agent goes to a Lava block.
    \item The DoorKey environment consists of a grid where the agent, a door, and a key are placed. The objective for the agent is to pick up the key, use it to unlock the door, and then navigate to a specified goal location, typically marked in a different color.
    \item The CartPole v1 environment consists of a cart constrained to move along a frictionless track, together with a pole balanced upright on the cart. The agent can apply forces to the left or right sides of the cart, and the objective is to make the pole stay upright as long as possible. The episode ends when the pole falls down, when the cart moves off the track, or when max steps is reached.
\end{itemize}

Below we have included examples of what the environments look like. A detailed description of these environments are in the appendix. 

\begin{figure}[h]
    \centering
    \begin{subfigure}{0.25\linewidth}
        \centering
        \includegraphics[width=\linewidth]{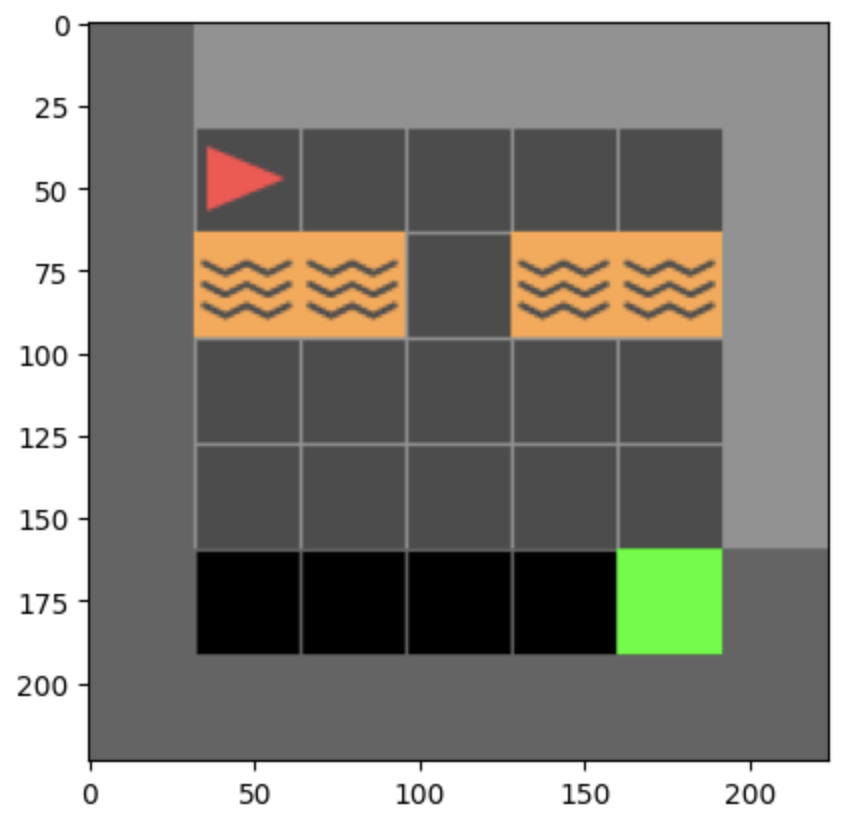}
    \caption{Crossing 7x7} 
    \label{fig:crossing}
    \end{subfigure} 
    \begin{subfigure}{0.25\linewidth}
        \centering
        \includegraphics[width=\linewidth]{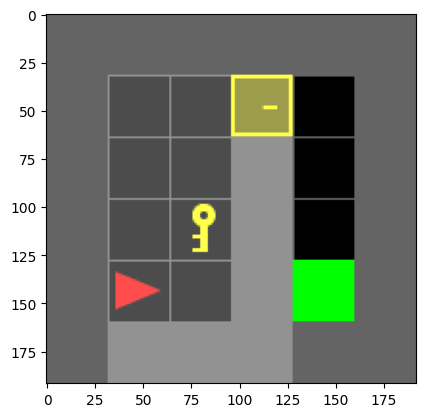}
        \caption{DoorKey 6x6}
        \label{fig:doorkey}
    \end{subfigure}
    \begin{subfigure}{0.25\linewidth}
        \centering
        \includegraphics[width=\linewidth]{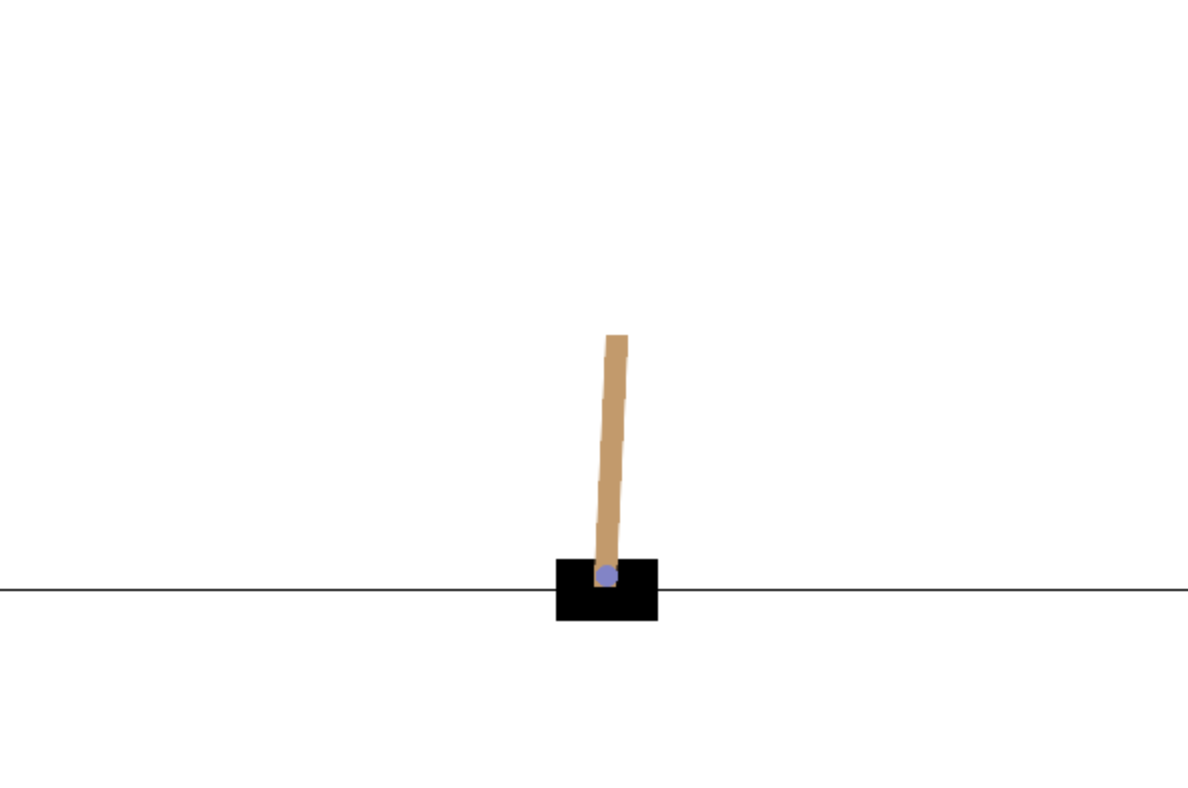}
    \caption{CartPole} 
    \label{fig:cartpole}
    \end{subfigure} 
\end{figure}

\subsection{Weak R3}

We first introduce a weaker version of R3 algorithm to illustrate how the idea of replay buffer for on-policy algorithm works.

\begin{algorithm}
\caption{Weak R3}
\begin{algorithmic}[1]
\State Initialization: PPO agent $\pi$, a cyclic buffer $\mathcal{B}$ with capacity $10$.
\While{steps $<$ max\_steps}
    \State Use agent $\pi$ to collect trajectory $D$; put $D$ into replay buffer $\mathcal{B}$ if $D$ succeeds.
    \State Train $\pi$ using $D$.
    \If{$\mathcal{B}$ is not empty}
        \State Randomly select $D'$ from $\mathcal{B}$ and use $D'$ to train $\pi$.
        \State If its fit is too low, then drop $D'$ from $\mathcal{B}$.
    \EndIf
\EndWhile
\end{algorithmic}
\end{algorithm}

Several remarks are in order. 
\begin{enumerate}
    \item By \emph{use a trajectory $D$ to train a PPO agent $\pi_\theta$}, we mean the following:
The trajectory $D$ is a set of data $\br{(s_i, a_i, p_i, r_i)}_{i = 1}^N$ such that $p_i = \pi_{\text{old}}(a_i | s_i)$, where $\pi_{\text{old}}$ is the agent that runs the trajectory.
Now let $L^{\PPO}_i(\pi, D)$ denote the PPO loss, namely,
\[
L^{\PPO}_i(\pi, D) = - \min\pr{\frac{\pi_\theta(a_i | s_i)}{\pi_{\hat{\theta}}(a_i | s_i)} \cdot \hat{A}_i, \clip\pr{\frac{\pi_\theta(a_i | s_i)}{\pi_{\hat{\theta}}(a_i | s_i)}, 1 - \epsilon, 1 + \epsilon} \hat{A}_i},
\]
where $\hat{\theta}$ is the original parameters and $\hat{A}_i$ is the advantage. Then the loss for weak R3 is
\[
L^{\RRR}(\pi, D) = \frac{1}{|S|} \cdot \sum_{i \in S} \pr{\frac{\pi_{\hat{\theta}}(a_i | s_i)}{p_i} \cdot L^{\PPO}_i(\pi, D) - e_{\theta} H(\cdot | s_i)},
\]
where $e_\theta$ is the entropy coefficient of agent $\pi_\theta$ and $S = \br{1 \le i \le N : \pi_{\hat{\theta}}(a_i | s_i) / p_i < \sigma}$.
Here $\sigma$ is a hyperparameter which represents the threshold above which we discard a term.

\item In line 7, the \emph{fit} of $D$ with respect to $\pi$ is defined as $\fit_\pi(D) = |S| / N$.

\item If we let $\sigma$ approach infinity, no term would be discarded. However, then there will be a high variance that makes the training process unstable and less likely for the agent to converge. In other words, the introduction of an upper bound $\sigma$ significantly reduces the variance. In practice, we can take $\sigma = 2$.

\item The choice to only discard terms with ratios that are too high but not terms with ratios that are too low (close to 0) derives from the fact that only the high ratio terms would create a high variance and make the training process unstable, but not the low ratio terms.
Moreover, in the gradient descent step, a high ratio term would shift the parameters too much from the old policy so that the agent may not converge. However, a low ratio term would not have this effect. 
\end{enumerate}

Our experiments showed that the performance of weak R3 is already significantly better than PPO on the DoorKey Environment (and similarly expected for the other experiments, though not tested).

\subsection{R3}

Now we focus on the R3 algorithm, which is an on-policy algorithm in an environment with discrete action space and sparse rewards. Below, the exploiter $X$ will be an advantage actor-critic network using GAE \cite{schulman2018highdimensional}, while the initiator $I$ and explorers $E$ will just be ordinary convolutional neural networks (without critics). The hyperparameters are not optimally tuned, but the specific ones shown here work well in practice.

\begin{algorithm}
\caption{R3}
\begin{algorithmic}[1]
\State Initialization:
\State $\bullet$ Cyclic buffers $\mathcal{B}$ and $\mathcal{B}_{\text{large}}$, where $\mathcal{B}$ has capacity 10 and $\mathcal{B}_{\text{large}}$ has capacity 20.
\State $\bullet$ One initiator $I$, with entropy coefficient $0.5$.
\State $\bullet$ Two explorers $E_1, E_2$.
Their entropy coefficients are 0.03 and 0.02 respectively.
\State $\bullet$ One exploiter $X$, with entropy coefficient $0.01$.

\While{steps $<$ max\_steps}
    \If{no reward has been collected yet}
        \State Run initiator $I$ to collect trajectory $D$.
        \State Use $D$ to train both $I$ and $X$.
    \Else
        \If{$\mathcal{B}$ is not empty}
            \State Run exploiter $X$ to collect trajectory $D$.
            \State Use $D$ to train $X$.
            \If{$D$ did not succeed}
                \State Use a randomly selected trajectory $D'$ from $\mathcal{B}$ to train $X$.
                \State Drop $D'$ from $\mathcal{B}$ if its fit is too low.
            \EndIf
        \Else
            \State Randomly select explorer $E \in \br{E_1, E_2}$ and run $E$ to collect trajectory $D$.
            \State Use $D$ to train both $E$ and $X$.
            \State Randomly select a trajectory $D'$ from $\mathcal{B}_{\text{large}}$ to train $E_2$.
        \EndIf
    \EndIf
    \If{$D$ is successful}
        \State Add $D$ to both $\mathcal{B}$ and $\mathcal{B}_{\text{large}}$.
    \EndIf
\EndWhile
\end{algorithmic}
\end{algorithm}

For clarity, we discuss the idea in the main training loop (starting from line 6). The whole training process consists of three phases: starting phase, exploration phase, and exploitation phase. 
\begin{itemize}
    \item There is an initiator in the starting phase, and its only task is to find the first successful data. Once a successful data is found, the initiator will not be involved in any future work. The initiator is set with a high entropy coefficient that sacrifices stability for more exploration. The reason why we set such a high entropy coefficient for the initiator is that we do not really care about the stability of the initiator: the only thing we care about is how fast the initiator can obtain the first successful trajectory.
    \item Once the initiator collects a successful trajectory, we put it into both replay buffers and proceed to the exploitation phase. In this phase the exploiter collects data and trains itself. The hyperparameters of the exploiter PPO agent should be roughly the same as a standard PPO agent. Additionally, during each training step, if the exploiter did not succeed in this trajectory, then it randomly picks a past successful trajectory from the replay buffer and trains based on it. If its fit (defined in the previous subsection) is lower than a certain threshold $\vartheta$ ($\vartheta$ can be a function that varies with the length of replay buffer), then we simply drop this trajectory from the replay buffer. As per usual in the algorithm, if we collect a successful trajectory during this phase, we insert it into both replay buffers.
    \item Whenever the replay buffer becomes empty because all existing experiences have been discarded, we turn to the exploration phase. In this phase, there are two explorers whose initial parameters are copied from the exploiter at the start of exploration phase, and whose task is to explore until they find a successful trajectory to reenter the exploit phase. Additional, one of the two explorers (the one with smaller entropy coefficient) will learn from all past successful trajectories, not just the ones which fit its distribution. As per usual, if we collect a successful trajectory during this phase, we put it into both replay buffers and turn to the exploitation phase.
\end{itemize}

Some remarks:
\begin{enumerate}
    \item Both the initiator and the explorers do not use critic network (and hence are just plain CNNs).
    The reason is that their only task is to explore until they find a single successful trajectory, and a PPO agent without critic network can obtain its first successful trajectory faster than a PPO agent with a critic network, since it takes extra time to train the value network.
    \item Whenever the averaged successful rate is high enough (say $50\%$), we stop using the replay buffer, so that the R3 algorithm degenerates to PPO. The reason behind this choice is that now there are enough successful trajectories, so that the agent does not have to replay old ones to get new ones and any further replay may decrease the stability of the training process.
    \item We also experimented with replacing the initiator with a random decision maker that chooses actions randomly. Empirically this replacement has lowered the stability of the algorithm. One possible explanation of this phenomenon is that an PPO initiator with high entropy coefficient can produce probabilities higher than those of a uniform distribution, so that the importance sampling factors will be smaller, making training more stable.
\end{enumerate}

\begin{center}
    \includegraphics[width=0.6\textwidth]{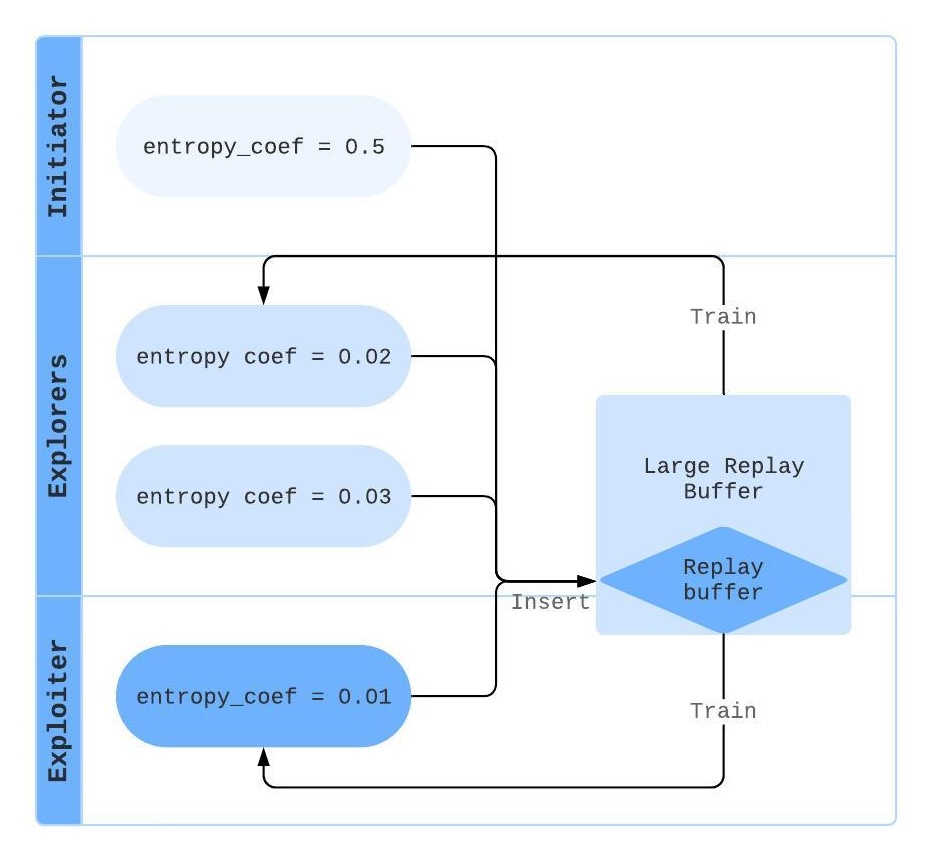}
    \captionof{figure}{Flowchart of R3}
    \label{fig:Flowchart}
\end{center}

\subsection{DR3}

The core idea of (weak) R3 can be generalized to a dense reward setting, which leads to the Dense R3 (DR3) algorithm.
Again, the hyperparameters are not optimally tuned, but the specific ones shown here work well in practice.

\begin{algorithm}
\caption{DR3}
\begin{algorithmic}[1]
\State Initialization: PPO agent $\pi$, a cyclic buffer $\mathcal{B}$ with capacity $20$.
\While{steps $<$ max\_steps}
    \State Use agent $\pi$ to collect trajectory $D$; put $D$ into replay buffer $\mathcal{B}$ if its reward is high.
    \State Train $\pi$ using $D$.
    \State Randomly select a $D'$ from $\mathcal{B}$ with high reward compared to $D$.
    \State If such a $D'$ exists, use $D'$ to train $\pi$; if the fit of $D'$ is too low, drop it from $\mathcal{B}$.
\EndWhile
\end{algorithmic}
\end{algorithm}

Again, several things need to be noted:
\begin{enumerate}
    \item By \emph{put $D$ into replay buffer $\mathcal{B}$ if its reward is high}, we mean either $r_D \ge \text{mean}(\mathcal{B})$ (when $\mathcal{B}$ is nonempty) or $r_D > \text{mean}(\mathcal{R}) + \text{std}(\mathcal{R})$ (when $\mathcal{B}$ is empty), where $r_D$ is the total reward of $D$ and $\mathcal{R}$ is the set of past (total) rewards.
    \item By \emph{randomly select a $D'$ from $\mathcal{B}$ with high reward compared to $D$}, we mean randomly select $D'$ from $\mathcal{B}$ with the property that $r_{D'} > r_D + \text{std}(\mathcal{B})$.
    \item Finally, when there is a maximum theoretical total reward $\overline{R}$, if the average total reward is above $\overline{R}/2$, we simply stop using the replay buffer, just like in R3. In this case, the DR3 algorithm also degenerates to PPO with the same reason of preserving the stability of the algorithm.
\end{enumerate}


\section{Results}
The advantage of R3 algorithm derives from the stability of the PPO agent and the sample efficiency of the replay buffer. Therefore, R3 performs especially well at tasks where rewards are difficult to obtain.

\subsection{Results in Crossing Env}
We tested R3 against PPO and DDQN on CrossingEnv. From the results, we can see that R3 underperforms DDQN on CrossingEnv. We think that the reason behind this phenomenon is that on this environment, on-policy algorithms are intrinsically worse than off-policy algorithms, since a $Q$ function can quickly learns not to step on lava and where to go whenever it sees a lava block. Also, it is fairly easy to explore a successful trajectory, so the real advantage of R3 cannot be disclosed on this environment.

However, on CrossingEnv the R3 agent outperforms PPO (and the advantage becomes much clear when we increase the size of the map), so we can say R3 is the best on-policy algorithm on this environment.
Also, we did not tune the hyperparameters of R3, so the performance could potentially be better.

\begin{figure}[H]
    \centering
    \begin{subfigure}{0.4\textwidth}
        \centering
        \includegraphics[width=\linewidth]{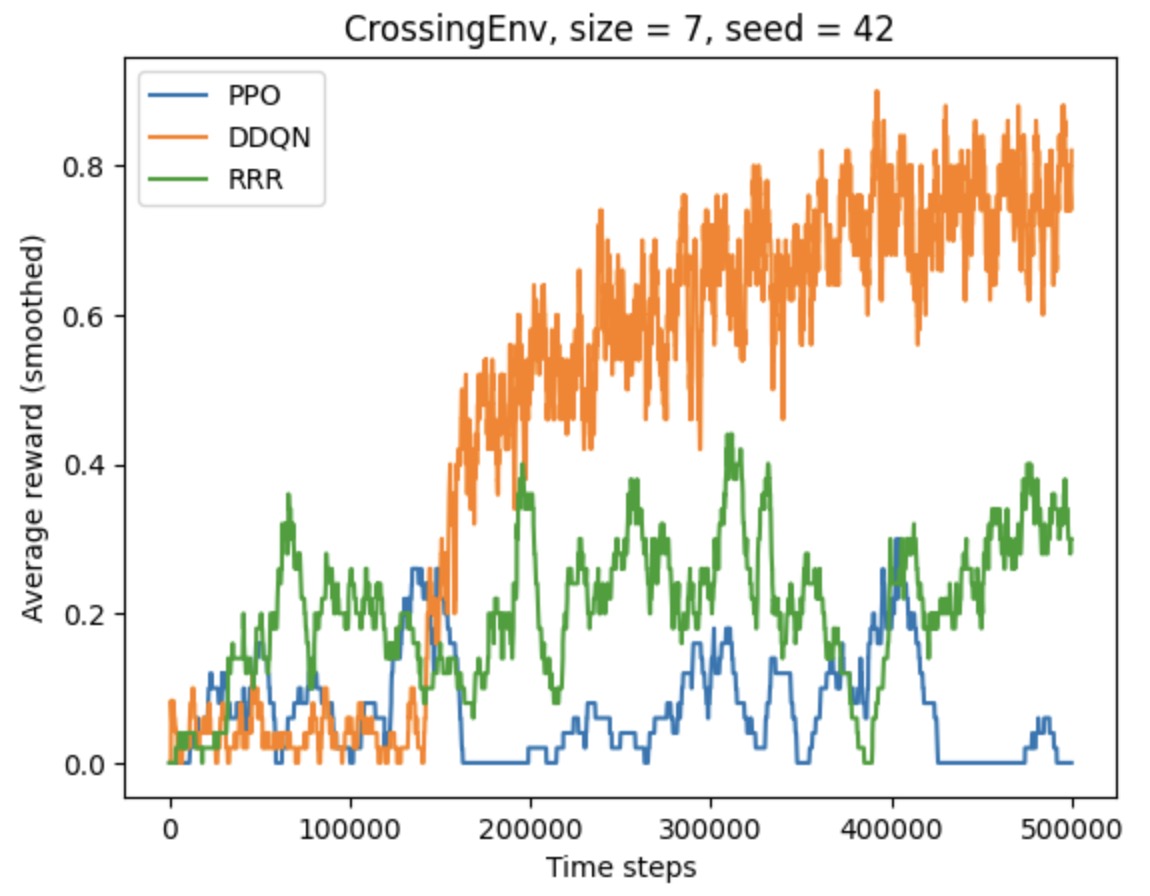}
        \caption{seed = 42}
    \end{subfigure}
    \begin{subfigure}{0.4\textwidth}
        \centering
        \includegraphics[width=\linewidth]{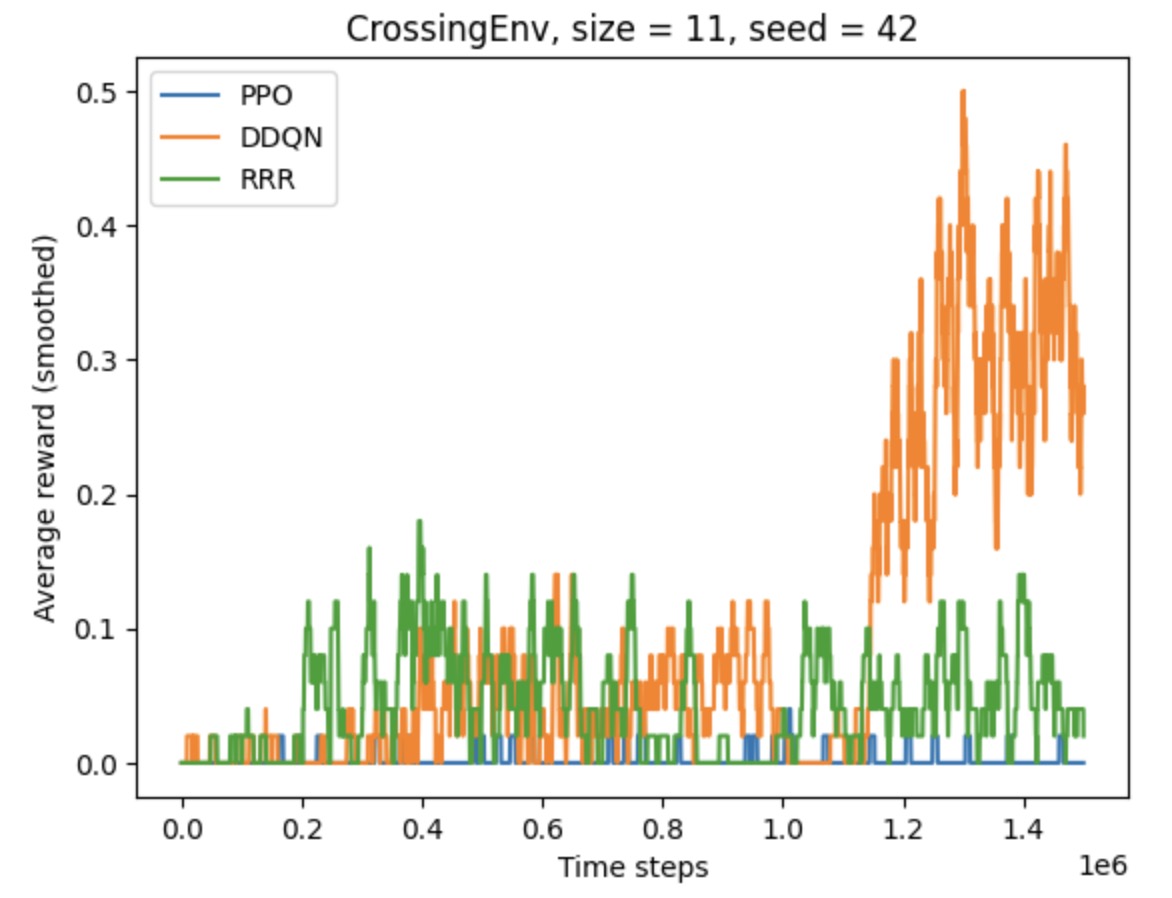}
        \caption{seed = 42}
    \end{subfigure}
    \hfill
    
    \caption{Performance of R3, PPO, and DDQN agent on Crossing environment}
    \label{fig:grid1}
\end{figure}

\begin{figure}[H]
    \centering
    \begin{subfigure}{0.25\textwidth}
        \centering
        \includegraphics[width=\linewidth]{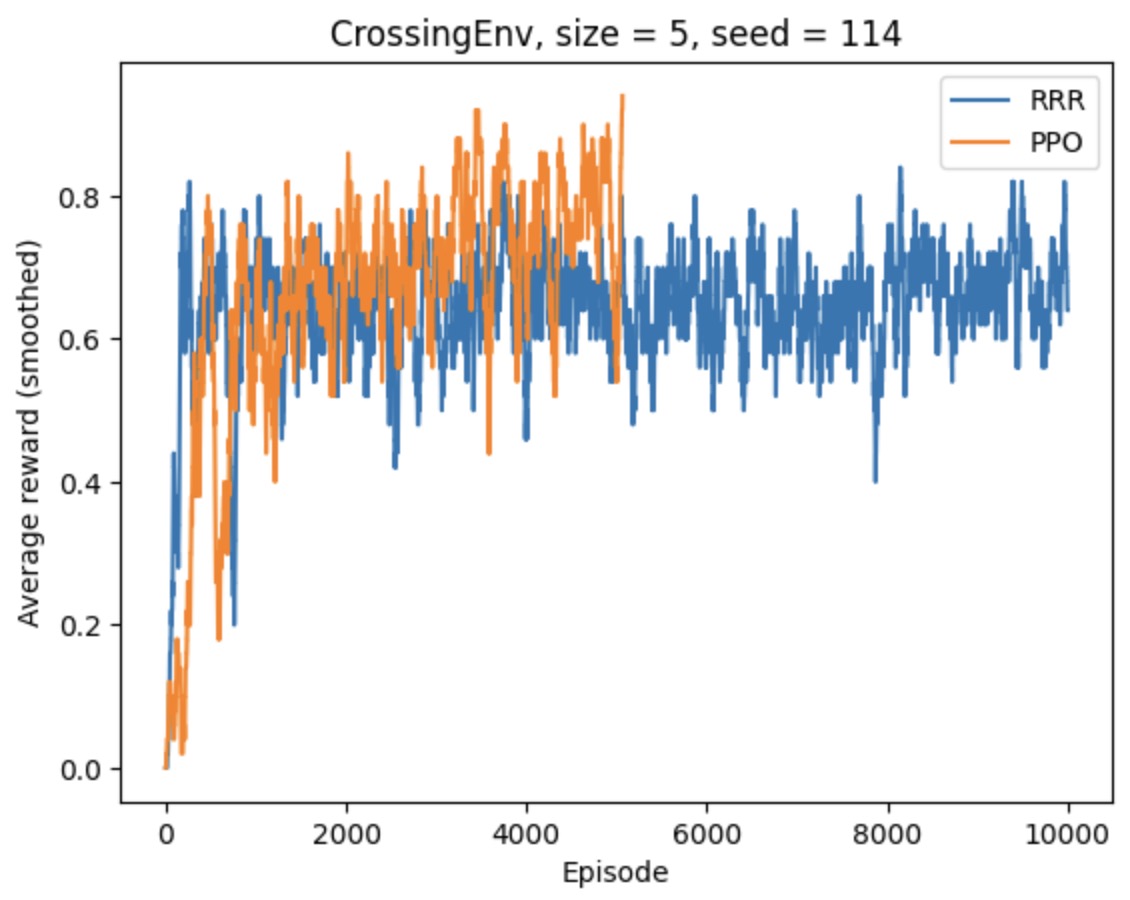}
        \caption{5x5, seed = 114}
    \end{subfigure}
    \hfill
    \begin{subfigure}{0.25\textwidth}
        \centering
        \includegraphics[width=\linewidth]{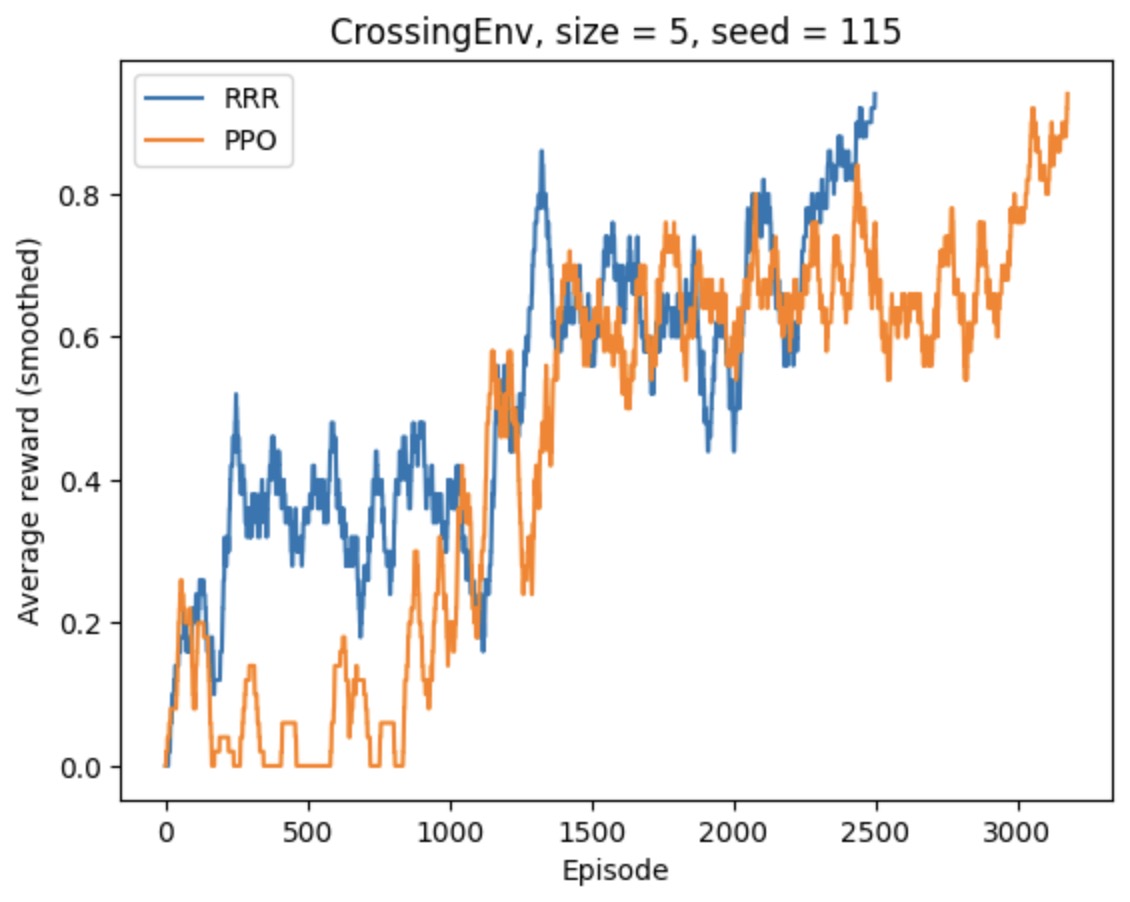}
        \caption{5x5, seed = 115}
    \end{subfigure}
    \hfill
    \begin{subfigure}{0.25\textwidth}
        \centering
        \includegraphics[width=\linewidth]{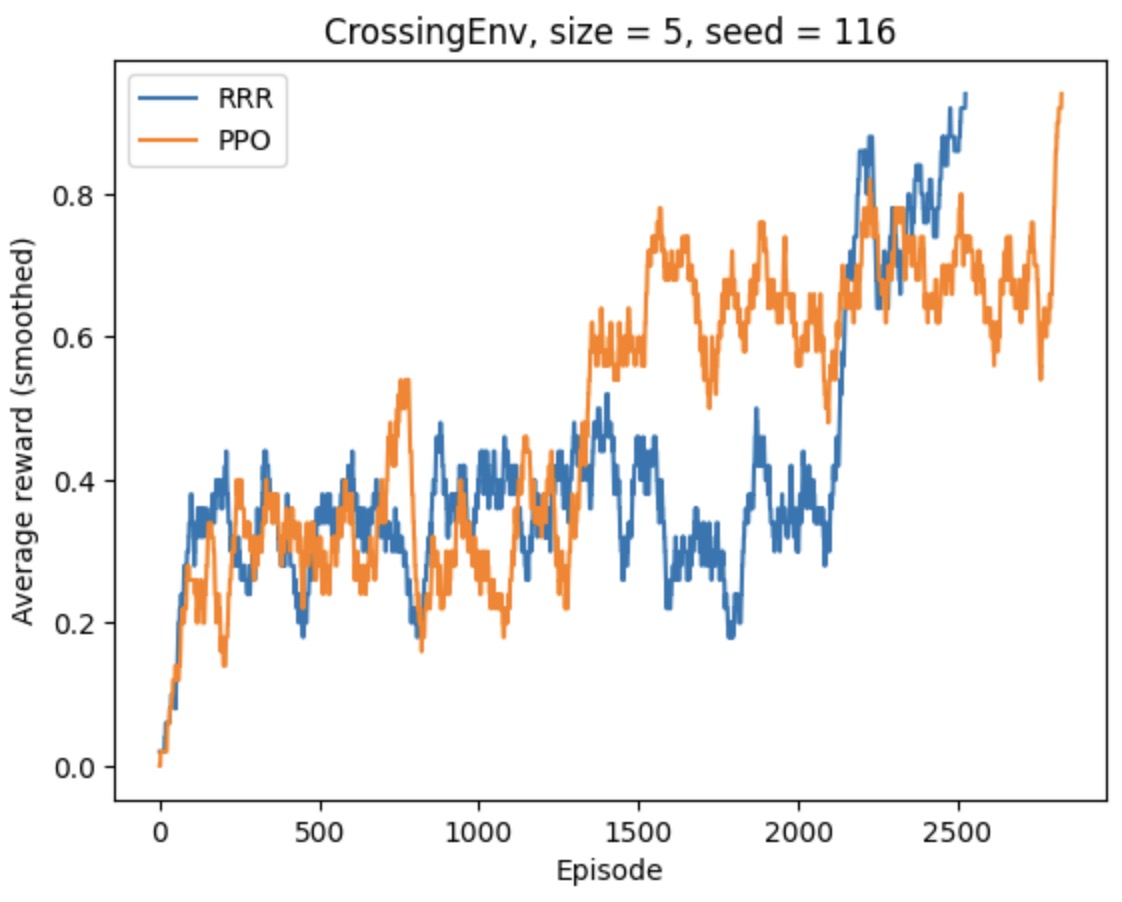}
        \caption{5x5, seed = 116}
    \end{subfigure}
    \\
    \begin{subfigure}{0.25\textwidth}
        \centering
        \includegraphics[width=\linewidth]{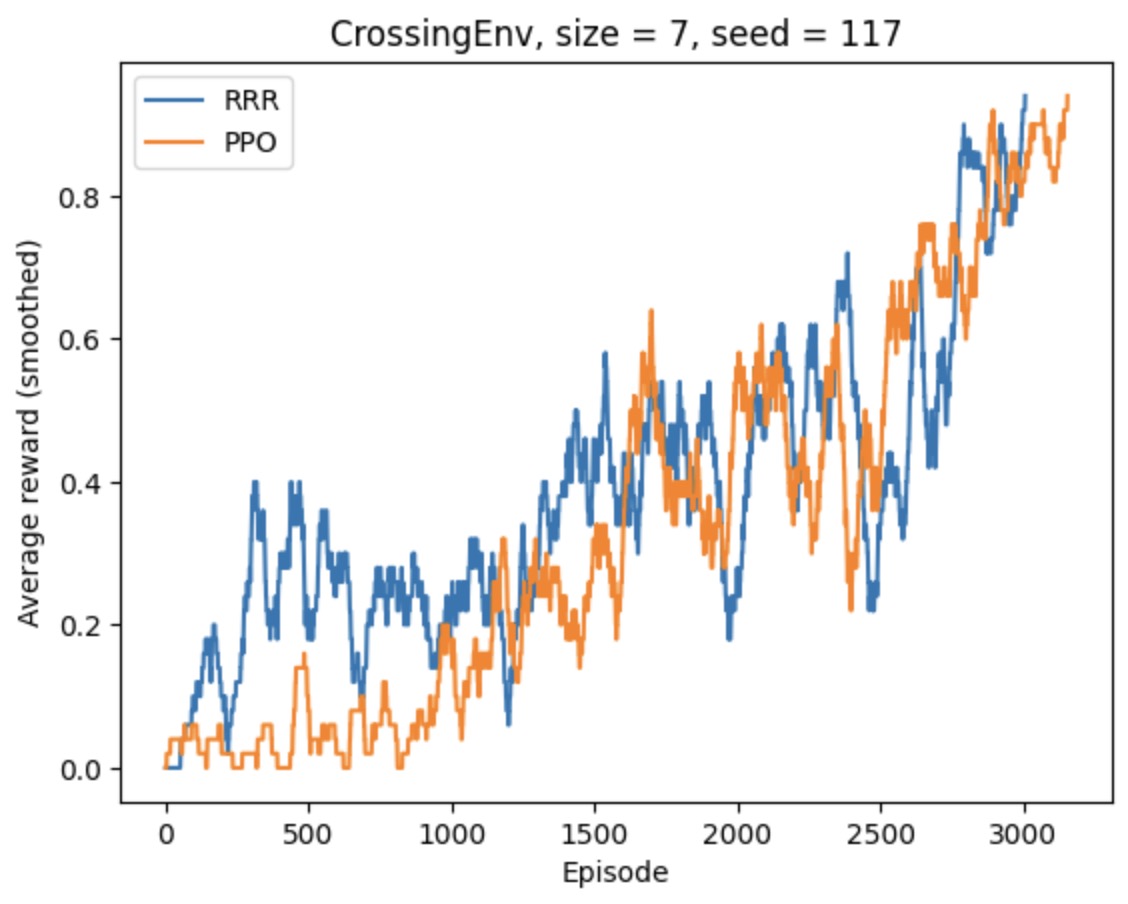}
        \caption{7x7, seed = 117}
    \end{subfigure}
    \hfill
    \begin{subfigure}{0.25\textwidth}
        \centering
        \includegraphics[width=\linewidth]{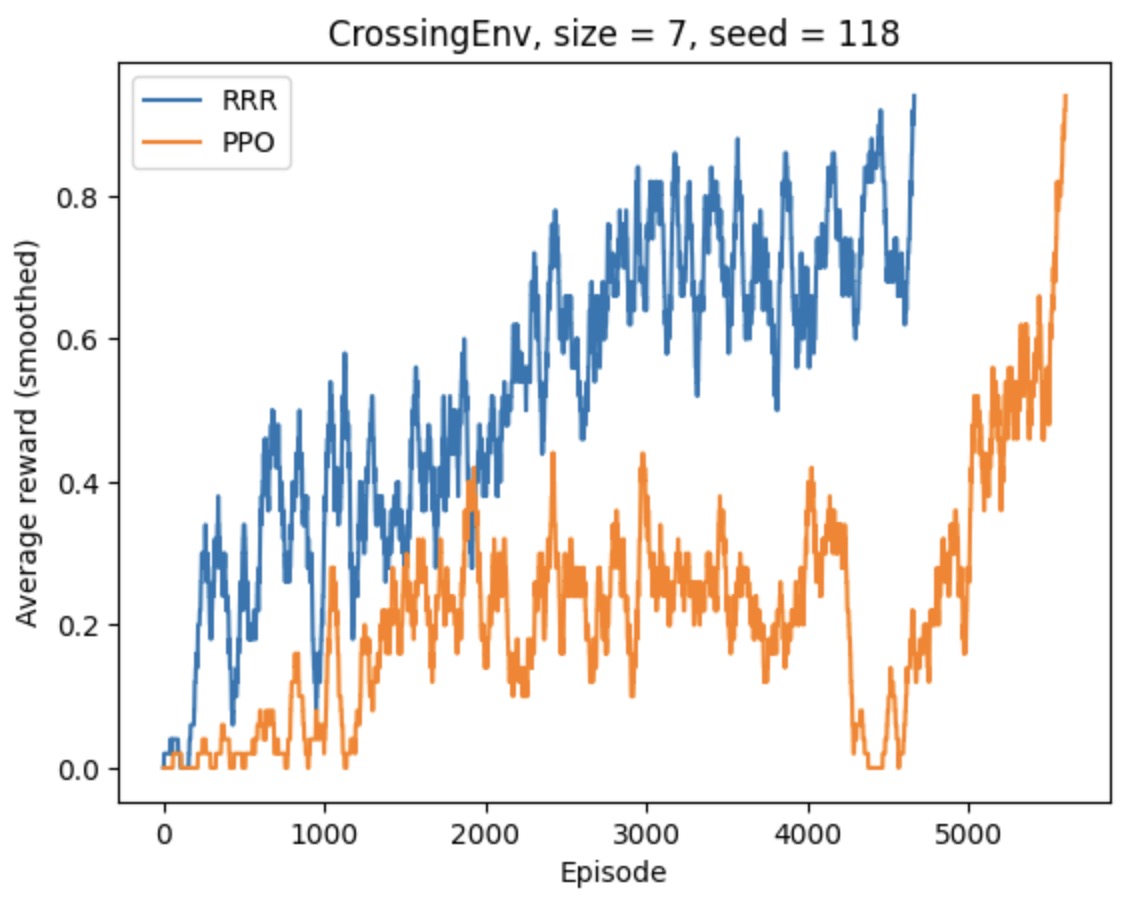}
        \caption{7x7, seed = 118}
    \end{subfigure}
    \hfill
    \begin{subfigure}{0.25\textwidth}
        \centering
        \includegraphics[width=\linewidth]{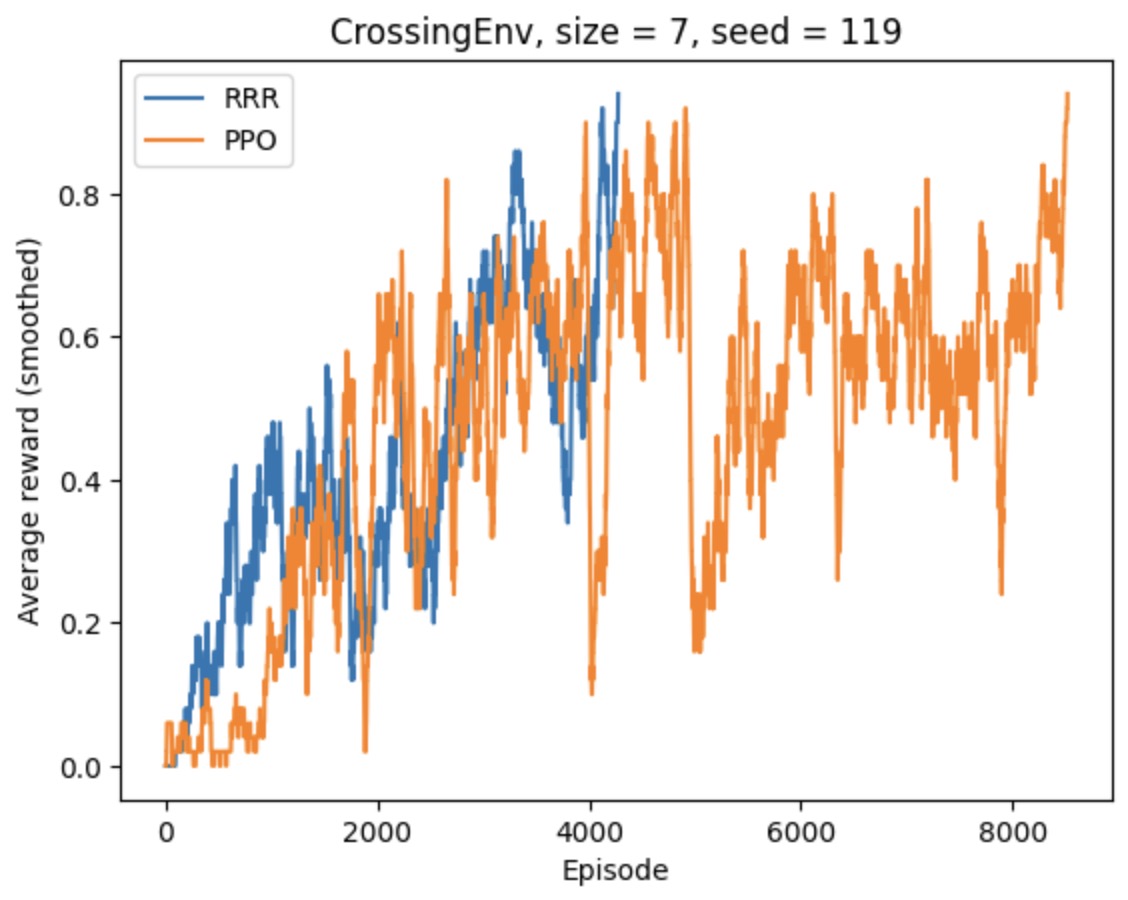}
        \caption{7x7, seed = 119}
    \end{subfigure}
    \\
    \begin{subfigure}{0.25\textwidth}
        \centering
        \includegraphics[width=\linewidth]{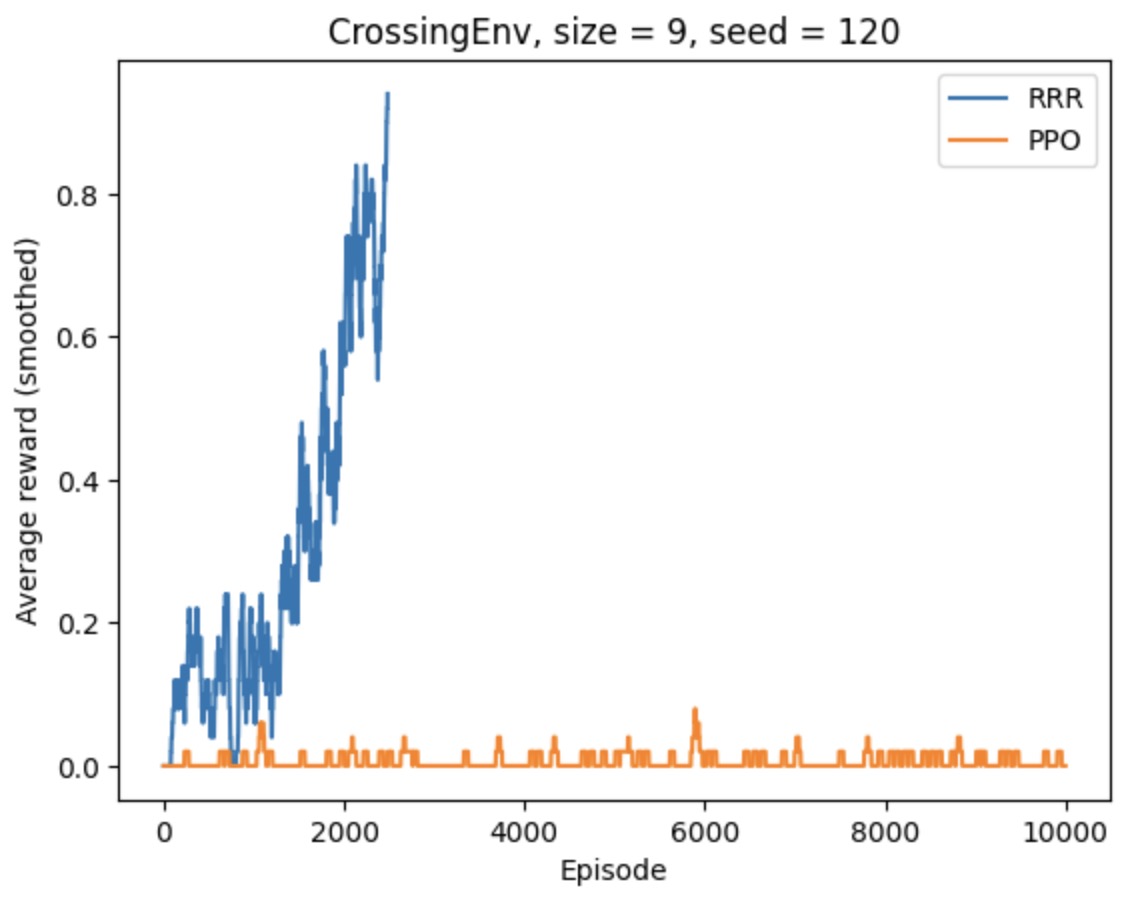}
        \caption{9x9, seed = 120}
    \end{subfigure}
    \hfill
    \begin{subfigure}{0.25\textwidth}
        \centering
        \includegraphics[width=\linewidth]{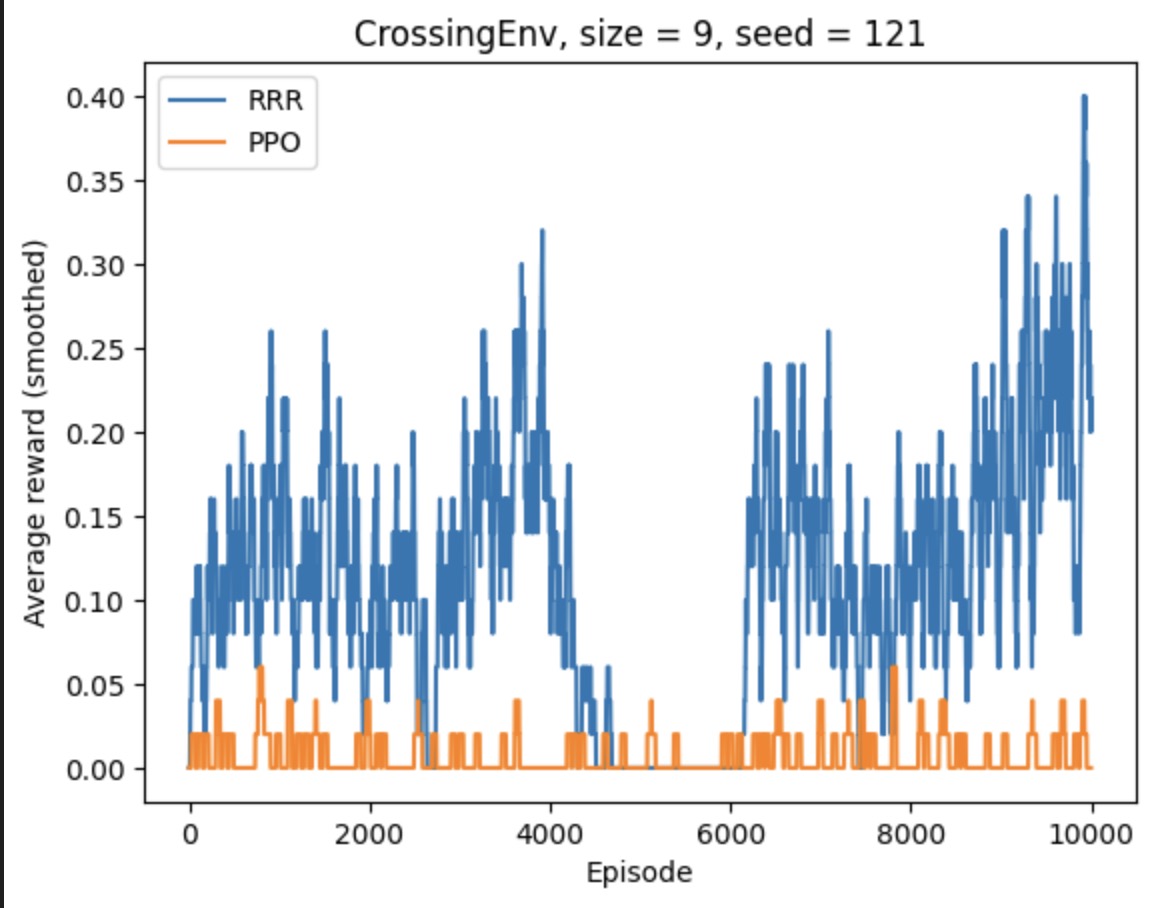}
        \caption{9x9, seed = 121}
    \end{subfigure}
    \hfill
    \begin{subfigure}{0.25\textwidth}
        \centering
        \includegraphics[width=\linewidth]{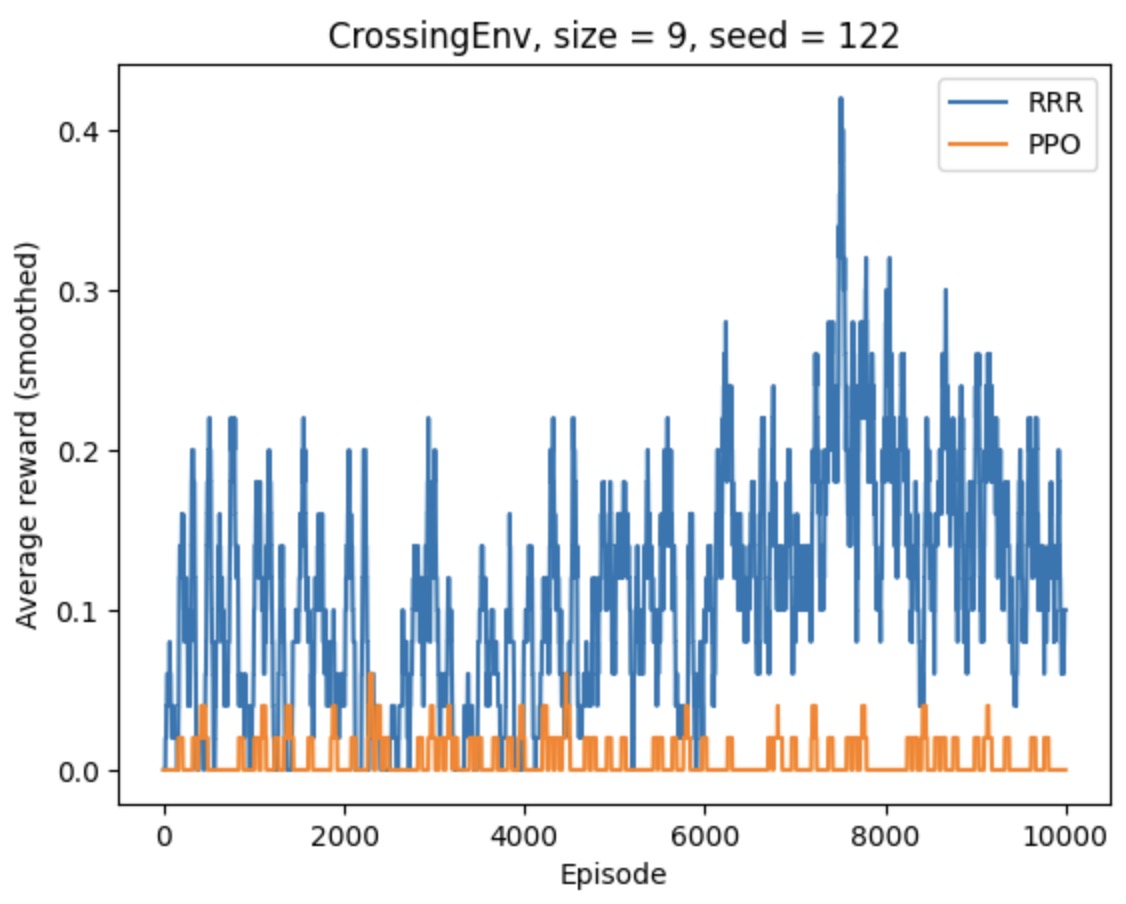}
        \caption{9x9, seed = 122}
    \end{subfigure}
    \caption{Performance of R3 vs PPO agent on Crossing environment}
    \label{fig:grid2}
\end{figure}

\subsection{Results in DoorKey Env}

The strength of R3 is most distinguishable in DoorKey Env. In this environment, it is rare to explore a successful trajectory. The agent needs to first pick up a key, and then use to open the door, and these two behaviors require two certain actions that are never used elsewhere.
From the results, we can see clearly that R3 outperforms both DDQN and PPO. From this we conclude that the more complicated the task is, the better R3 performs compared with DDQN and PPO. The advantage of R3 over DDQN and PPO increases with complexity of the task because is because R3 really knows how to use past successful trajectories to improve its current policy, so it can quickly learn from them.

An observation that is worth noting is that DDQN usually runs much longer than PPO and R3, and PPO and R3 takes about the same time to run.
For exanple, on DoorKeyEnv of size 8, the running time for DDQN was 4 hours 44 minutes, for PPO was 59 minutes, and for R3 was 1 hour 14 minutes.

\begin{figure}[H]
    \centering
    \begin{subfigure}{0.4\textwidth}
        \centering
        \includegraphics[width=\linewidth]{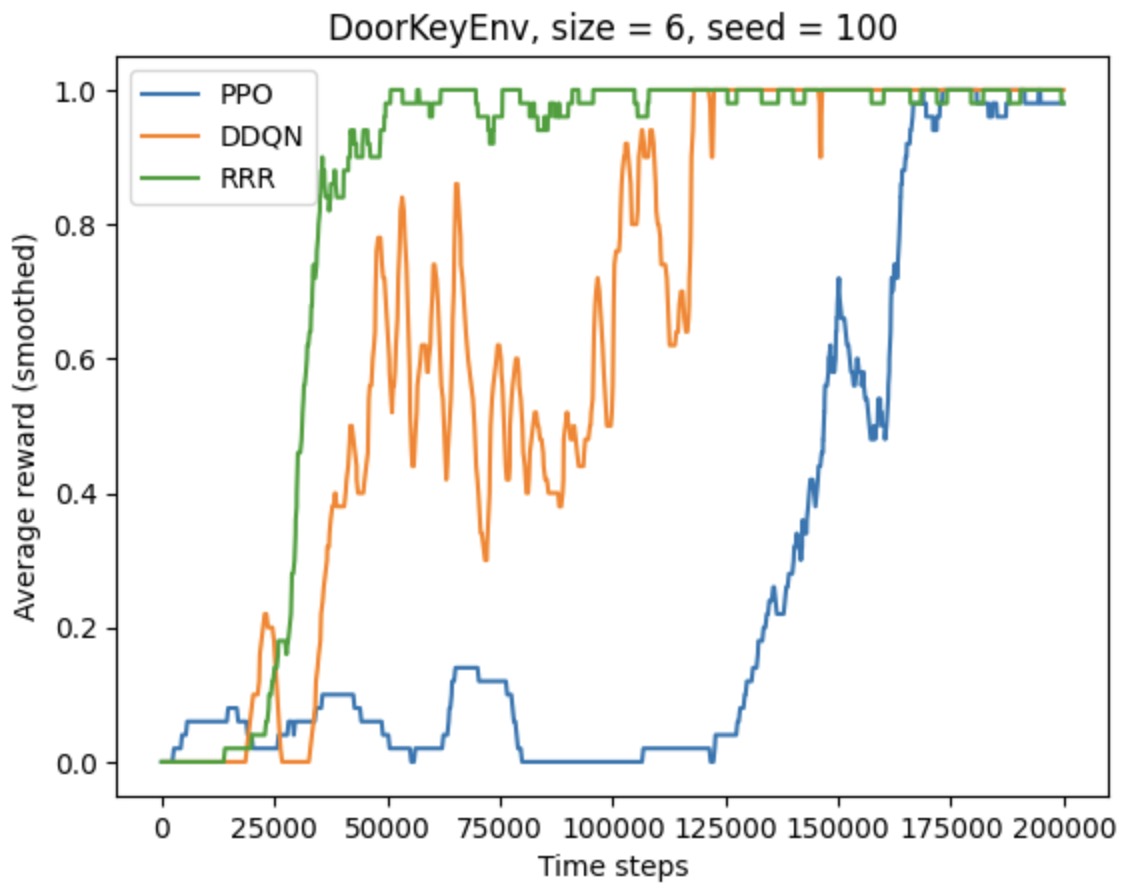}
        \caption{seed = 100}
    \end{subfigure}
    \begin{subfigure}{0.4\textwidth}
        \centering
        \includegraphics[width=\linewidth]{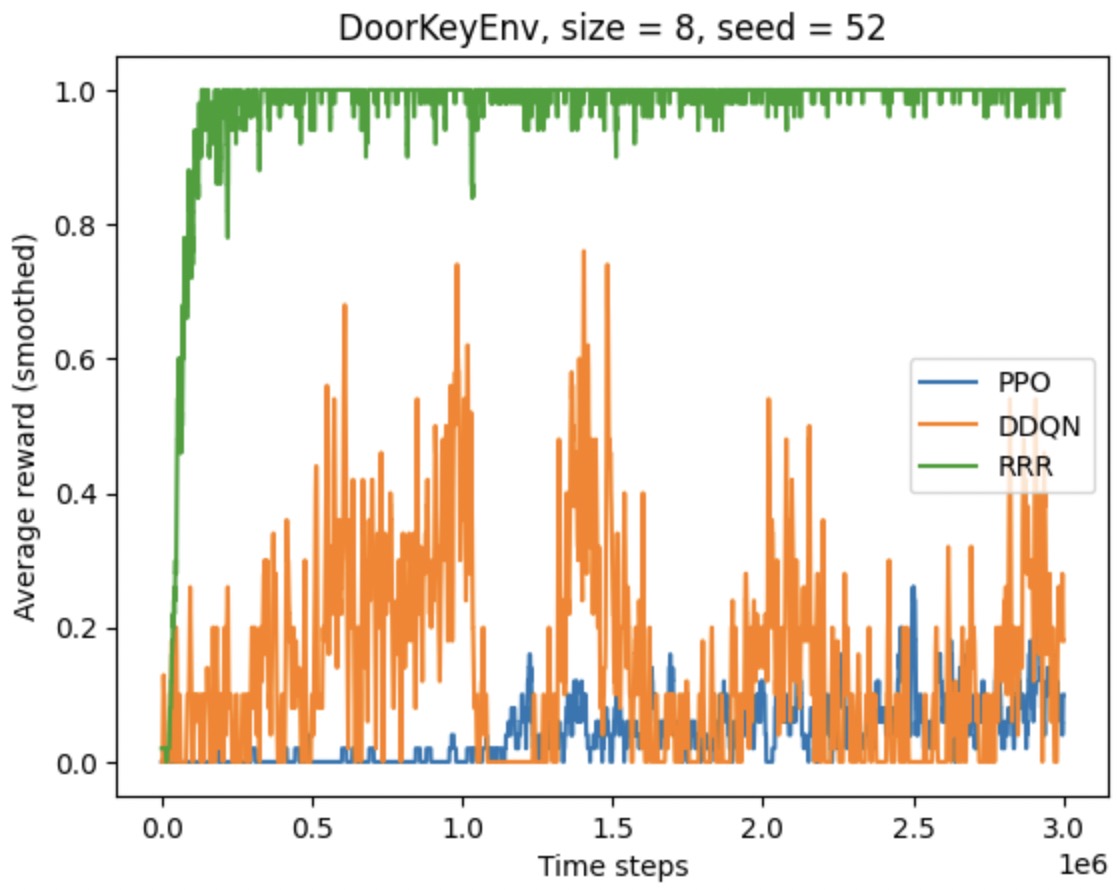}
        \caption{seed = 52}
    \end{subfigure}
    \hfill
    
    \caption{Performance of R3, PPO, and DDQN agent on DoorKey environment}
    \label{fig:grid3}
\end{figure}

\begin{figure}[H]
    \centering
    \begin{subfigure}{0.25\textwidth}
        \centering
        \includegraphics[width=\linewidth]{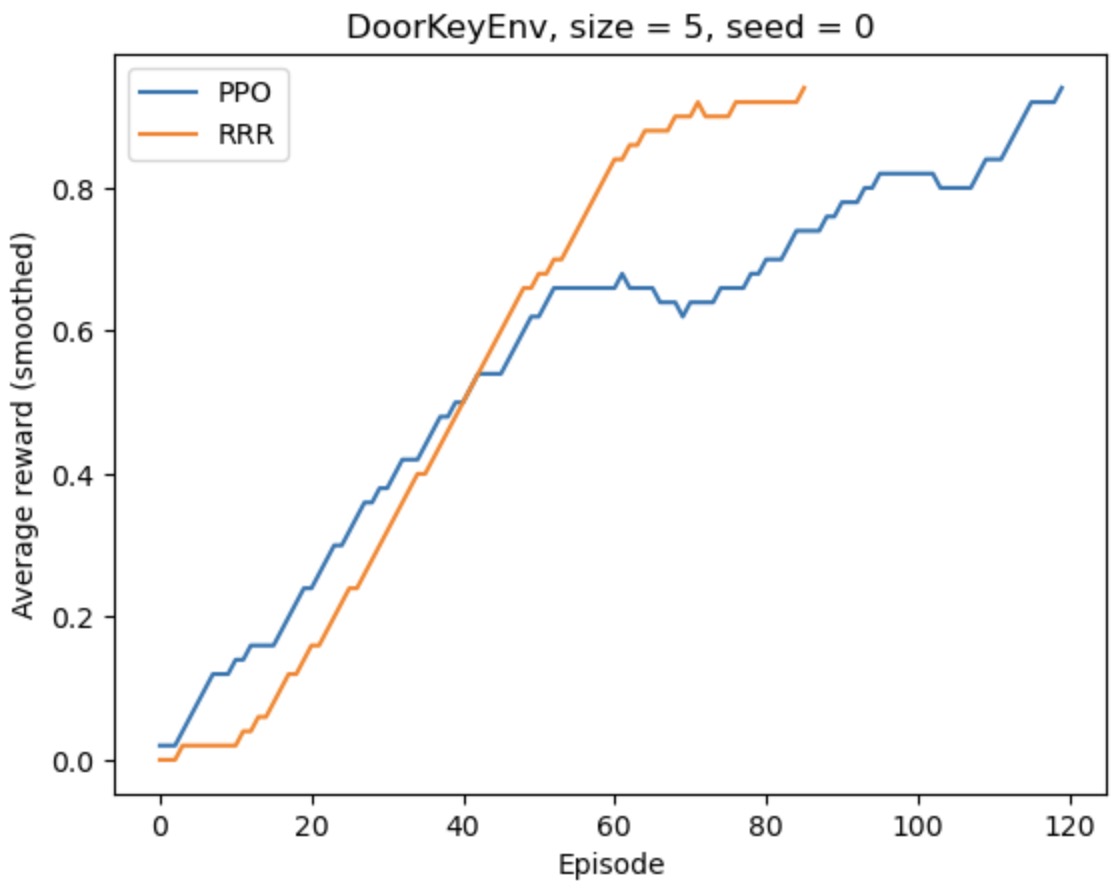}
        \caption{5x5, seed = 0}
    \end{subfigure}
    \hfill
    \begin{subfigure}{0.25\textwidth}
        \centering
        \includegraphics[width=\linewidth]{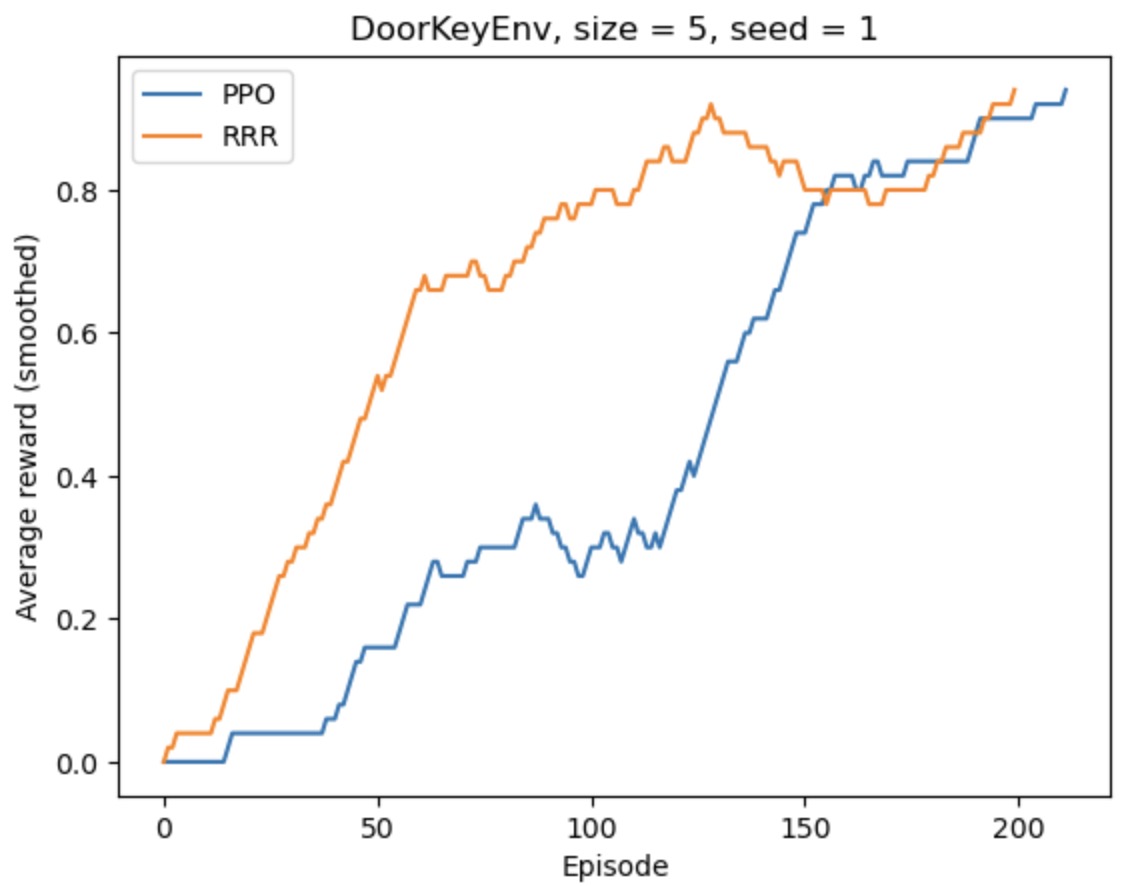}
        \caption{5x5, seed = 1}
    \end{subfigure}
    \hfill
    \begin{subfigure}{0.25\textwidth}
        \centering
        \includegraphics[width=\linewidth]{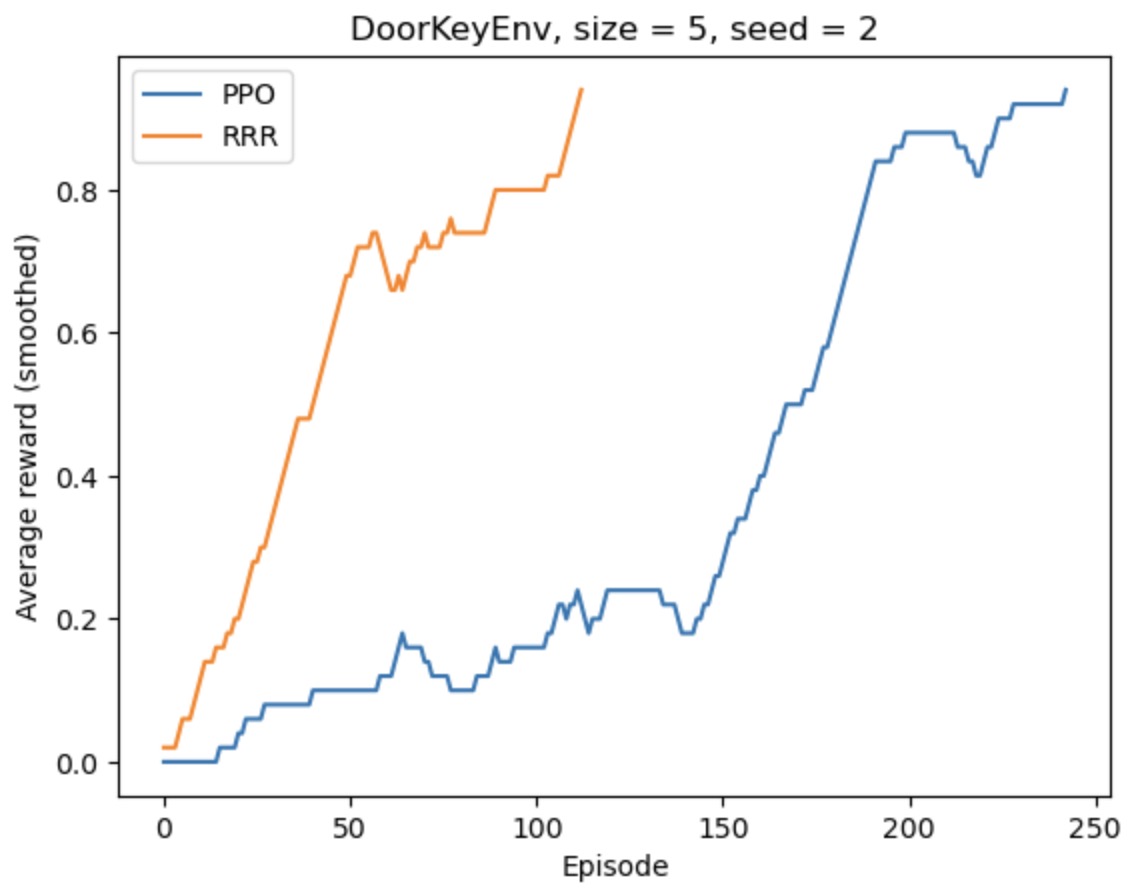}
        \caption{5x5, seed = 2}
    \end{subfigure}
    \\
    \begin{subfigure}{0.25\textwidth}
        \centering
        \includegraphics[width=\linewidth]{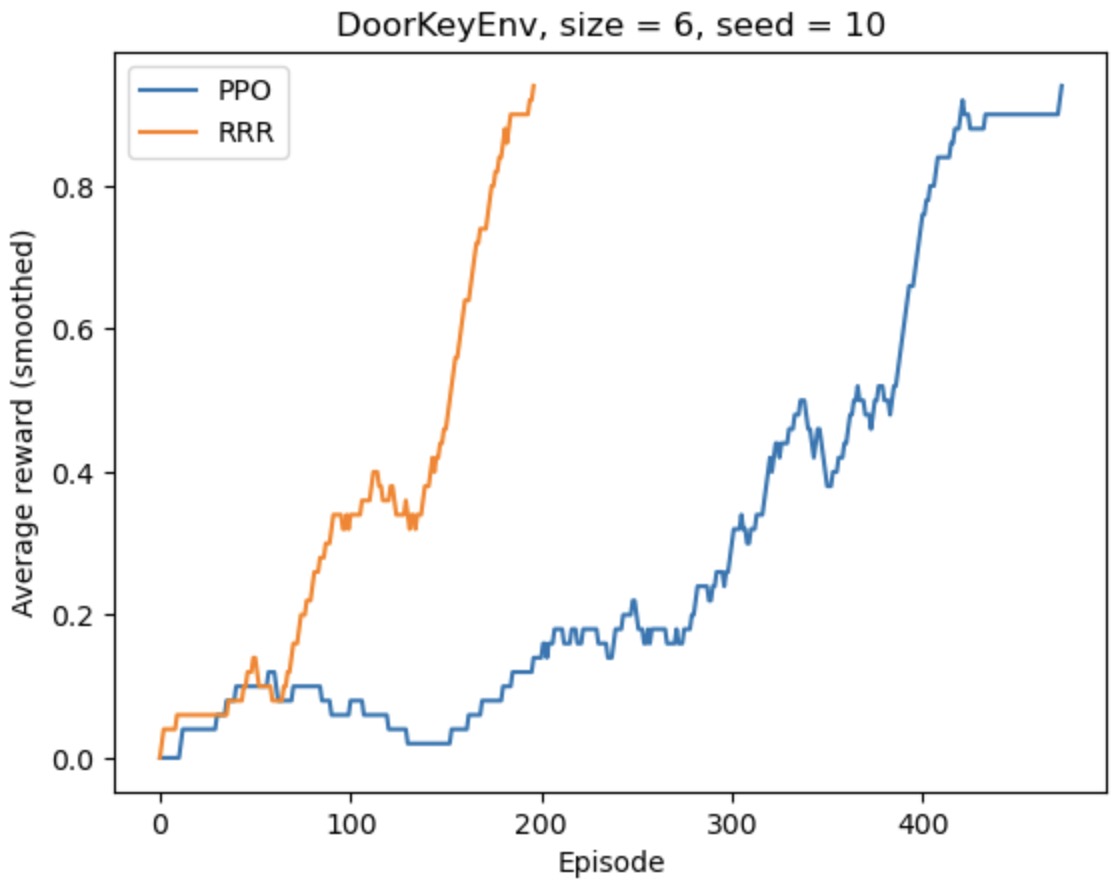}
        \caption{6x6, seed = 10}
    \end{subfigure}
    \hfill
    \begin{subfigure}{0.25\textwidth}
        \centering
        \includegraphics[width=\linewidth]{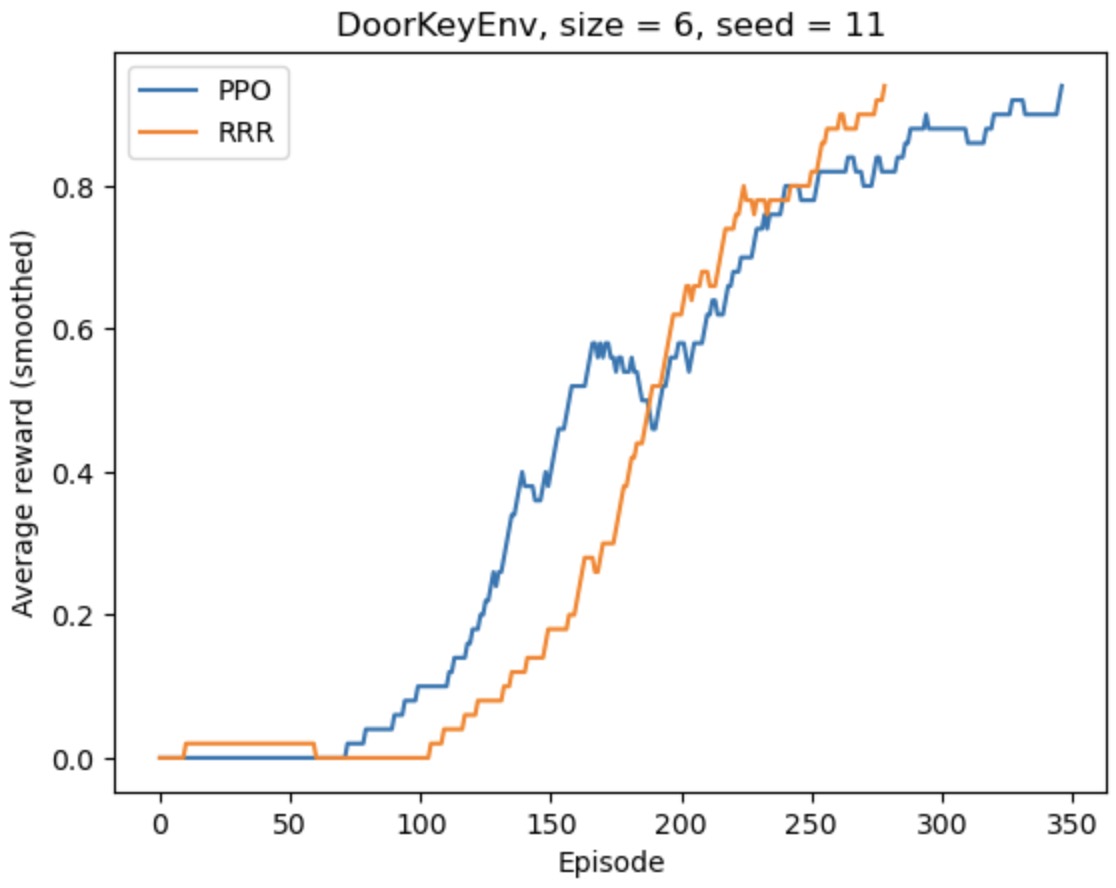}
        \caption{6x6, seed = 11}
    \end{subfigure}
    \hfill
    \begin{subfigure}{0.25\textwidth}
        \centering
        \includegraphics[width=\linewidth]{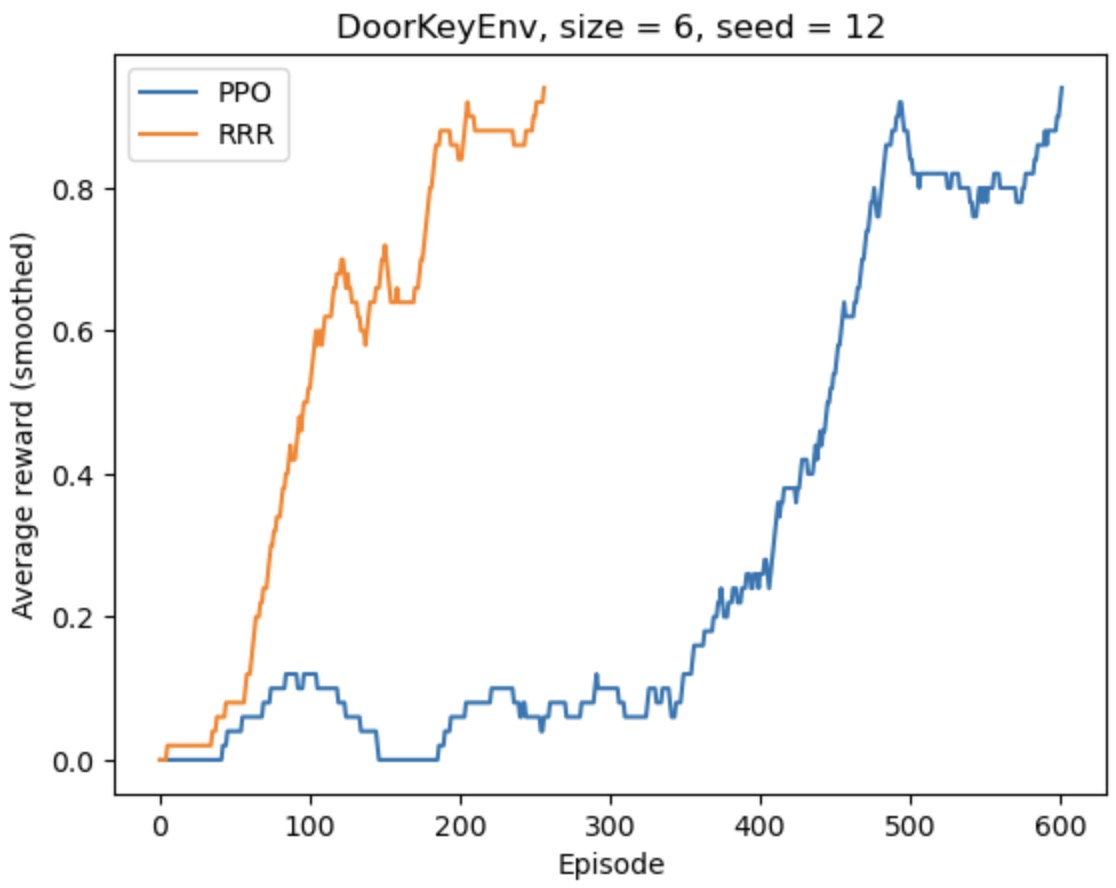}
        \caption{6x6, seed = 12}
    \end{subfigure}
    \\
    \begin{subfigure}{0.25\textwidth}
        \centering
        \includegraphics[width=\linewidth]{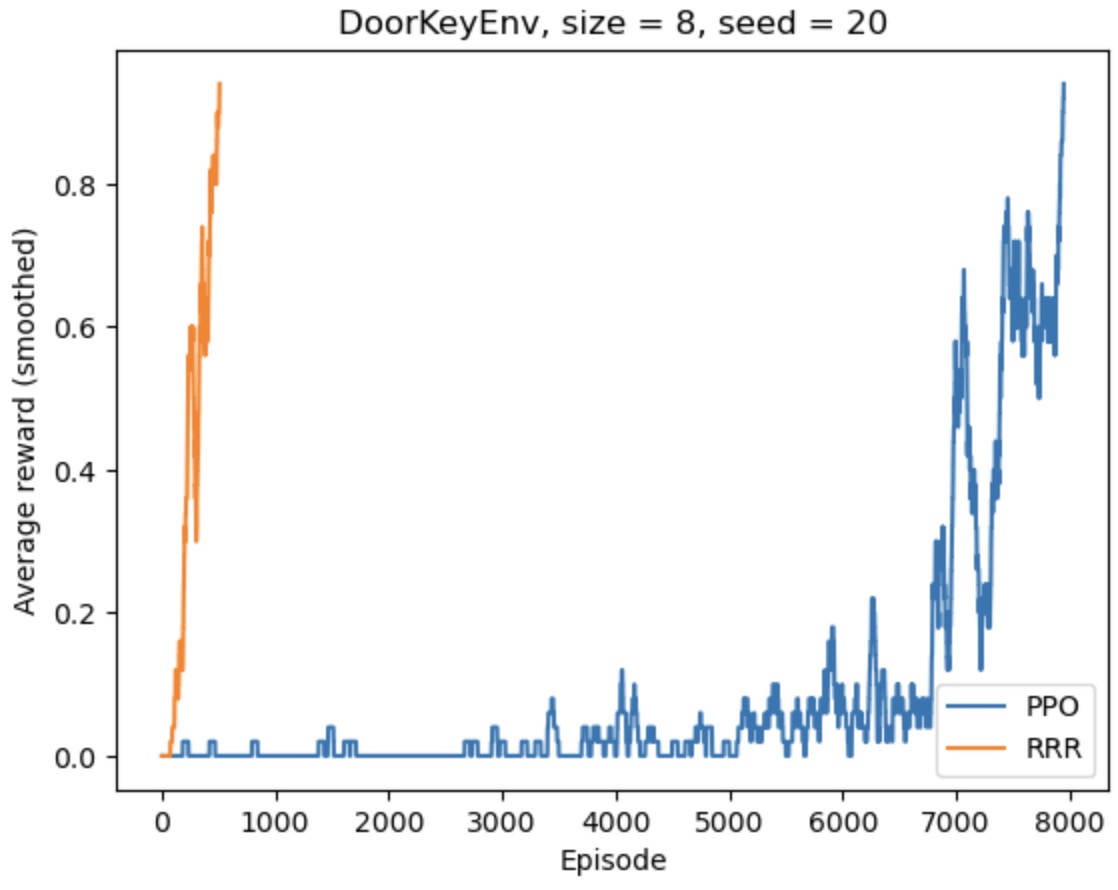}
        \caption{8x8, seed = 20}
    \end{subfigure}
    \hfill
    \begin{subfigure}{0.25\textwidth}
        \centering
        \includegraphics[width=\linewidth]{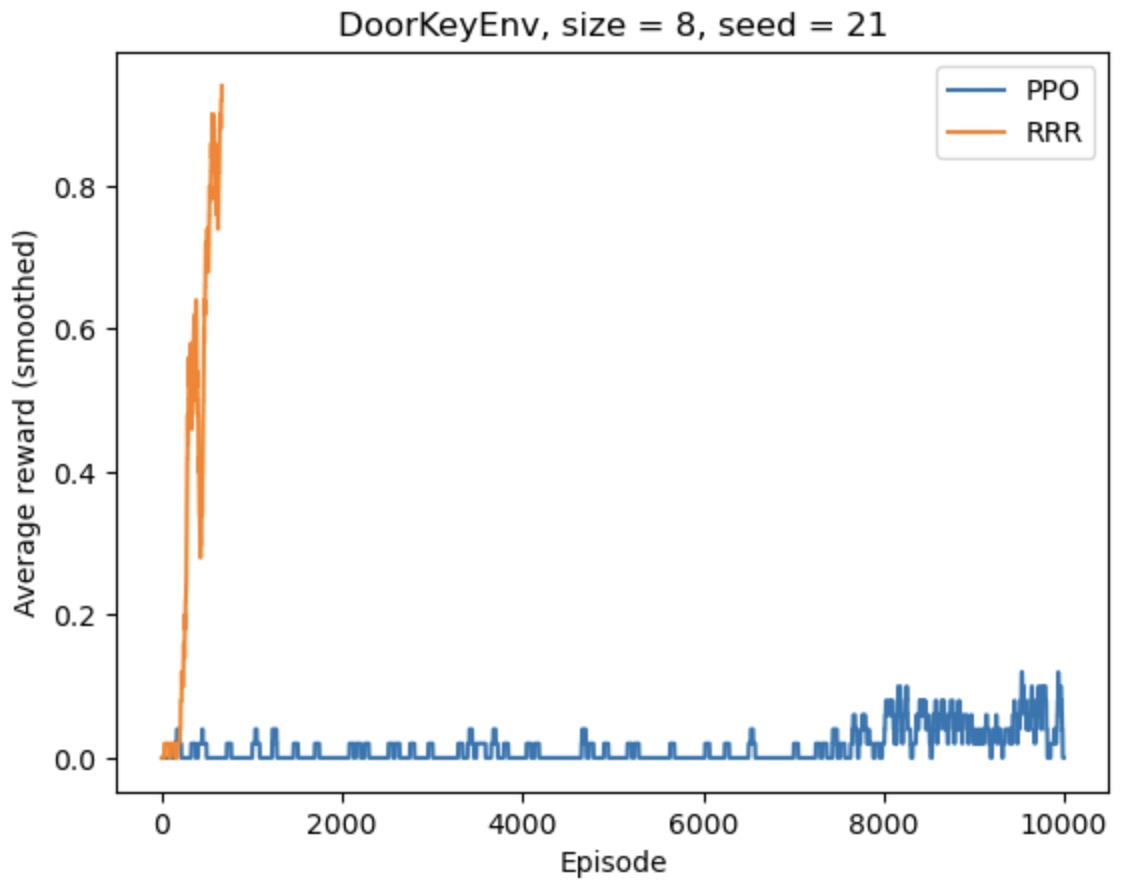}
        \caption{8x8, seed = 21}
    \end{subfigure}
    \hfill
    \begin{subfigure}{0.25\textwidth}
        \centering
        \includegraphics[width=\linewidth]{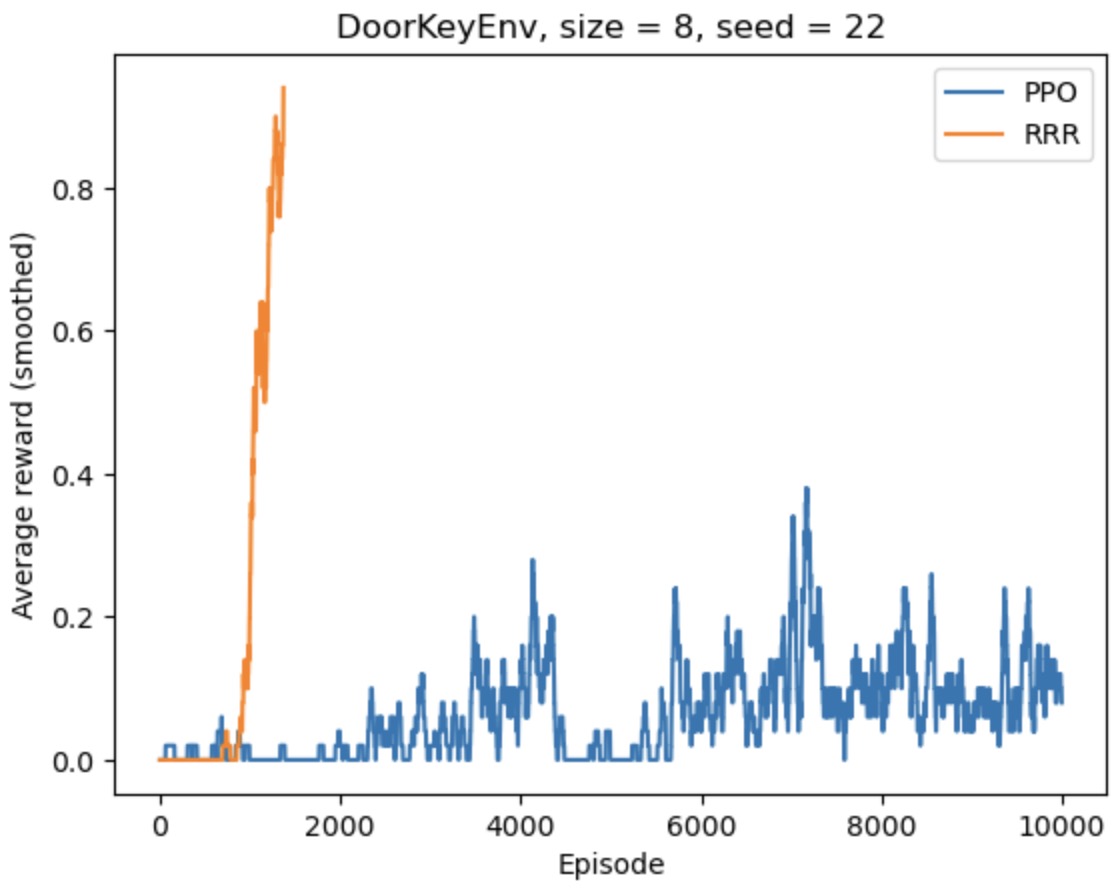}
        \caption{8x8, seed = 22}
    \end{subfigure}
    \caption{Performance of R3 vs PPO agent on DoorKey environment}
    \label{fig:grid4}
\end{figure}

\subsection{Results in CartPole Env}
We tested the DR3 agent with the baseline PPO agent in the Cartpole gym environments. Below are the results across 9 random seeds. The $y$-axis represents smoothed reward with maximum steps = 3000 and the $x$-axis represents the number of episodes. From the graphs it is evident that the DR3 agent significantly outperforms the baseline PPO agent.

Overall, in all of tasks R3 (or DR3) outperforms PPO, and usually by a significant margin. As a result, we think in general R3 (or DR3) is more optimal than PPO when the action space is discrete.
(D)R3 may transform current reinforcement learning society.

\begin{figure}[h]
    \centering
    \begin{subfigure}{0.25\textwidth}
        \centering
        \includegraphics[width=\linewidth]{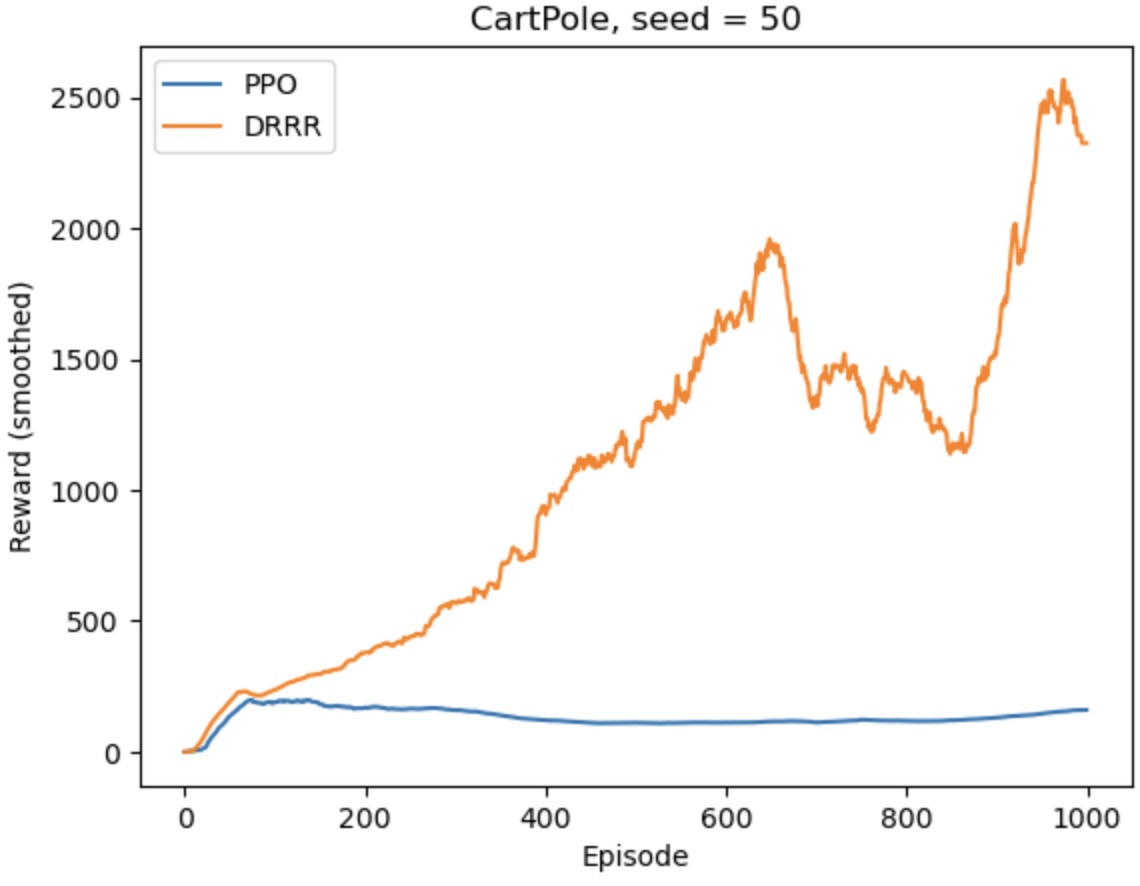}
        \caption{seed = 50}
        \label{fig:sub1}
    \end{subfigure}
    \hfill
    \begin{subfigure}{0.25\textwidth}
        \centering
        \includegraphics[width=\linewidth]{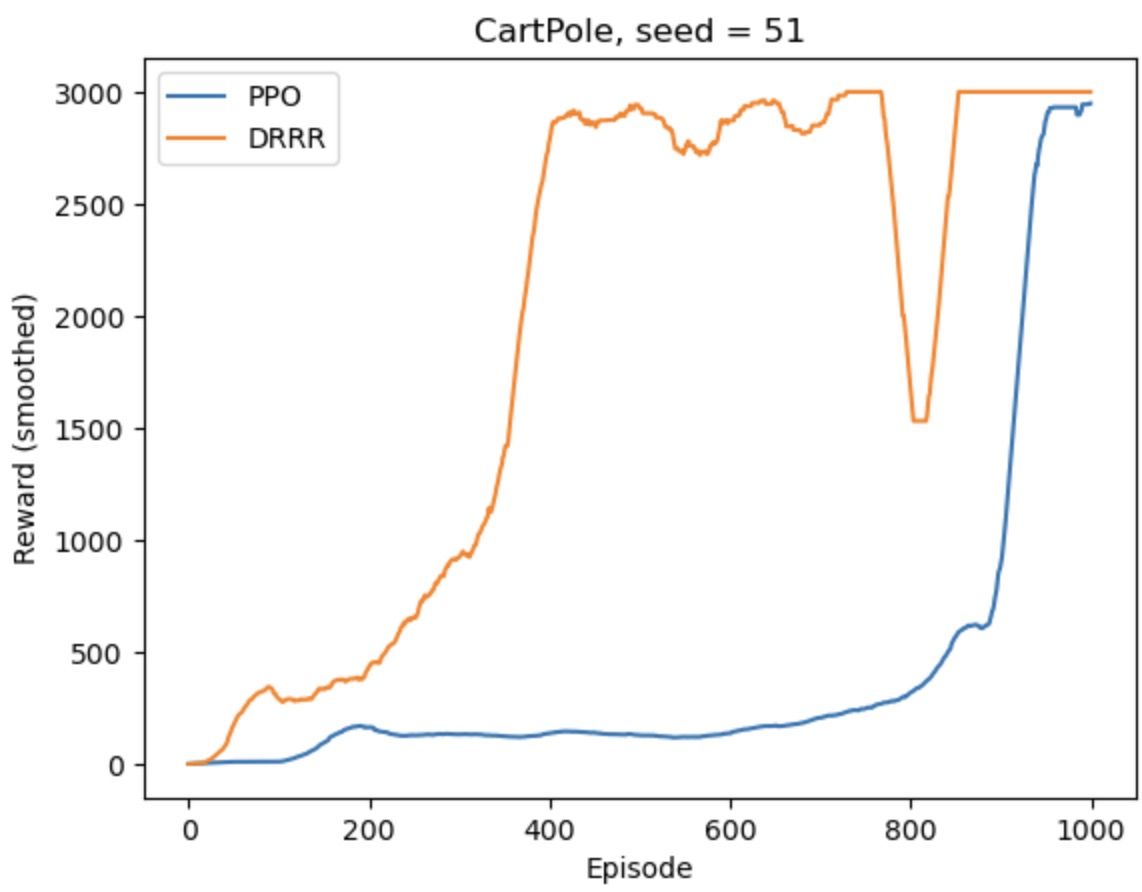}
        \caption{seed = 51}
        \label{fig:sub2}
    \end{subfigure}
    \hfill
    \begin{subfigure}{0.25\textwidth}
        \centering
        \includegraphics[width=\linewidth]{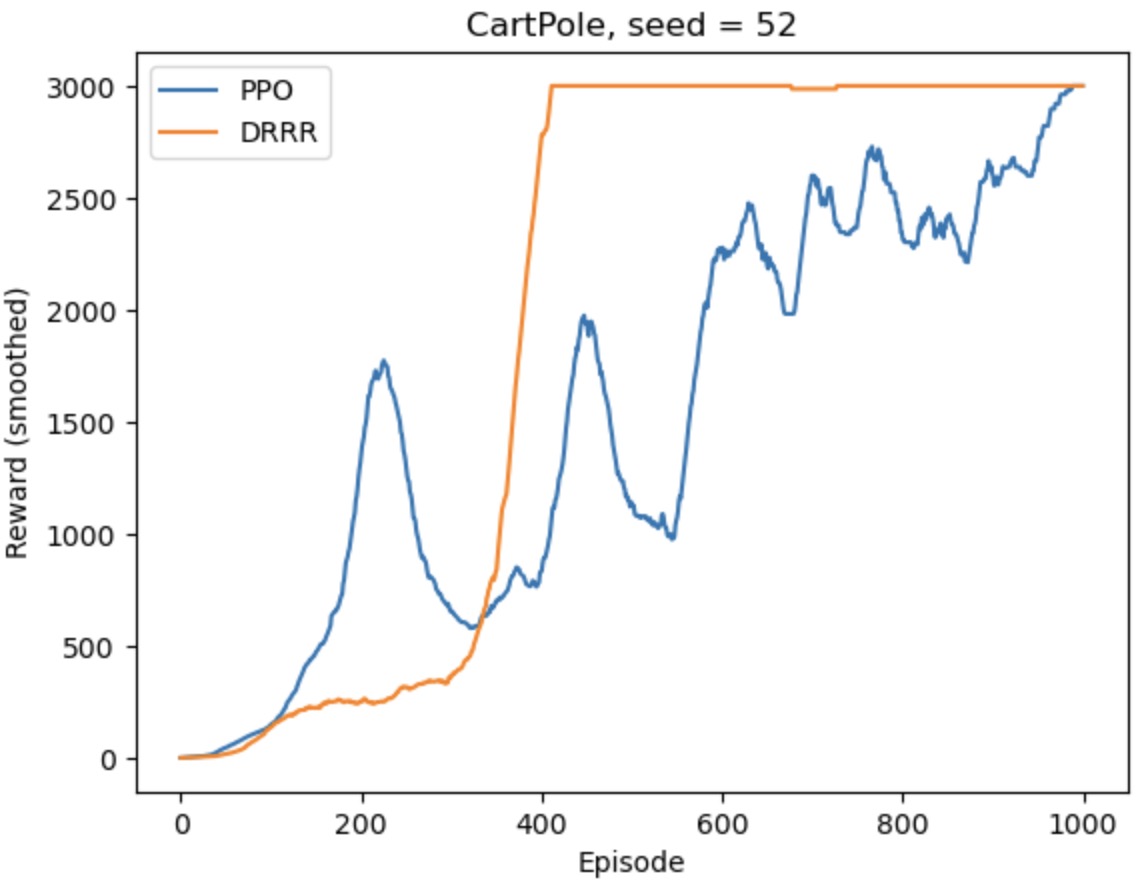}
        \caption{seed = 52}
        \label{fig:sub3}
    \end{subfigure}
    \\
    \begin{subfigure}{0.25\textwidth}
        \centering
        \includegraphics[width=\linewidth]{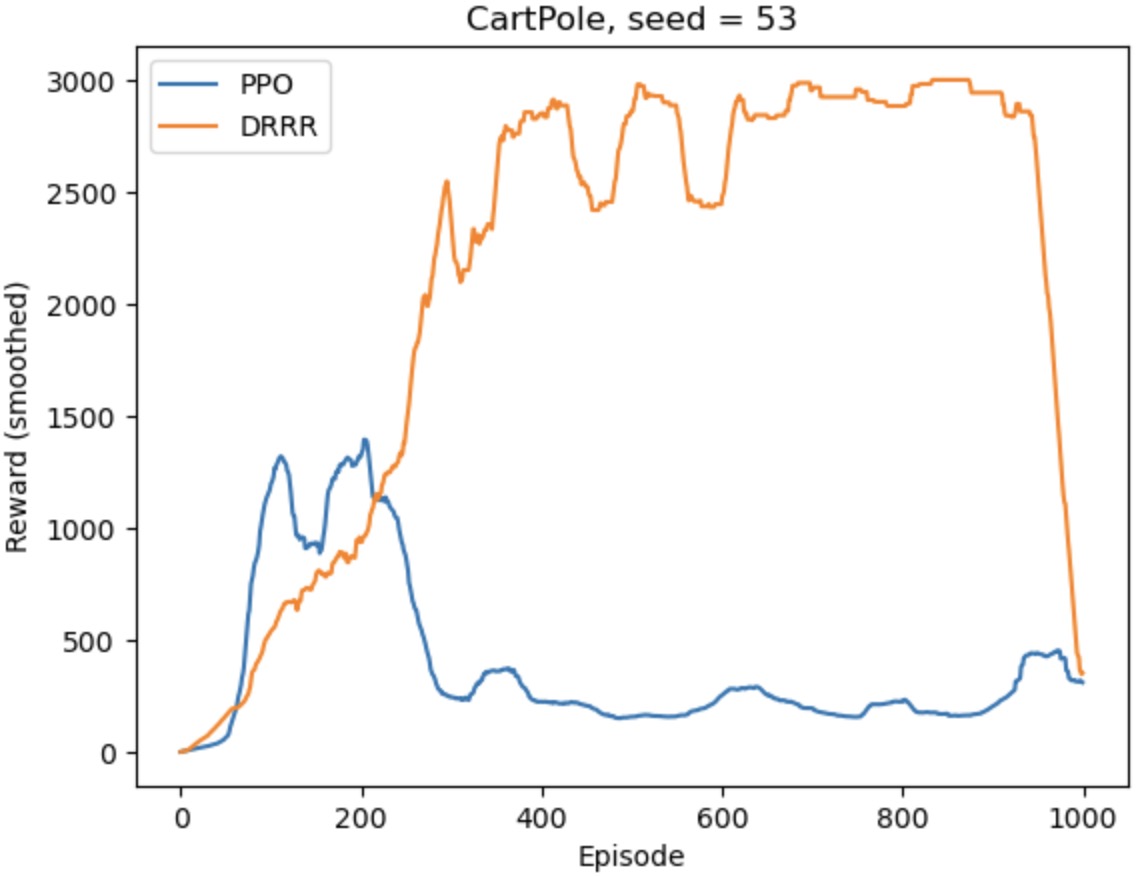}
        \caption{seed = 53}
        \label{fig:sub4}
    \end{subfigure}
    \hfill
    \begin{subfigure}{0.25\textwidth}
        \centering
        \includegraphics[width=\linewidth]{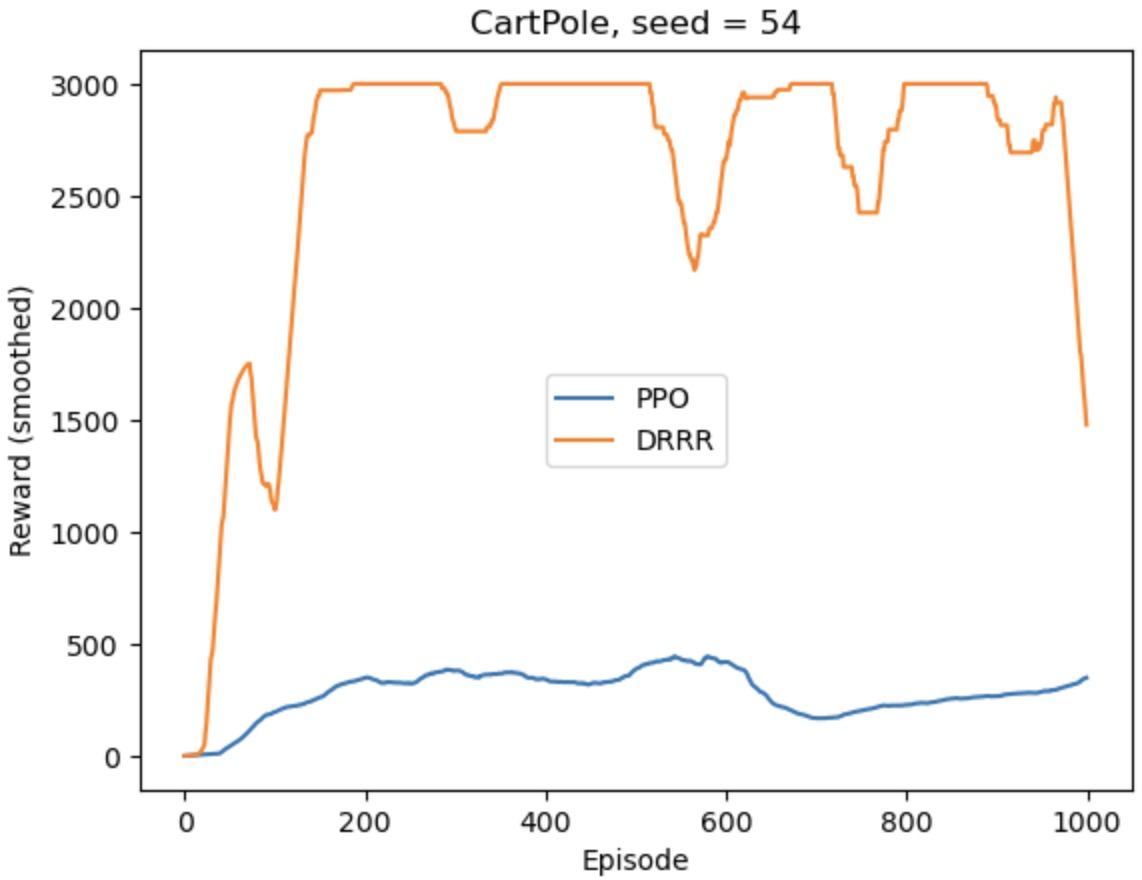}
        \caption{seed = 54}
        \label{fig:sub5}
    \end{subfigure}
    \hfill
    \begin{subfigure}{0.25\textwidth}
        \centering
        \includegraphics[width=\linewidth]{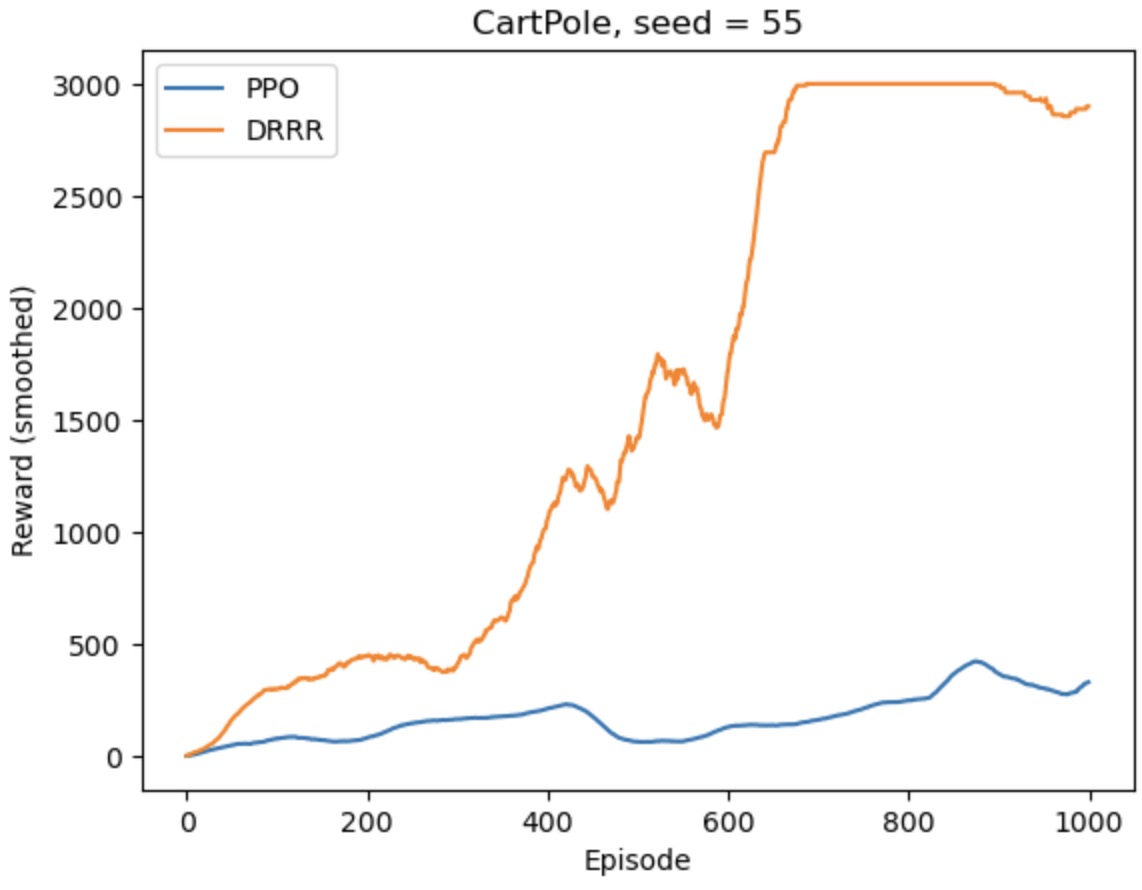}
        \caption{seed = 55}
        \label{fig:sub6}
    \end{subfigure}
    \\
    \begin{subfigure}{0.25\textwidth}
        \centering
        \includegraphics[width=\linewidth]{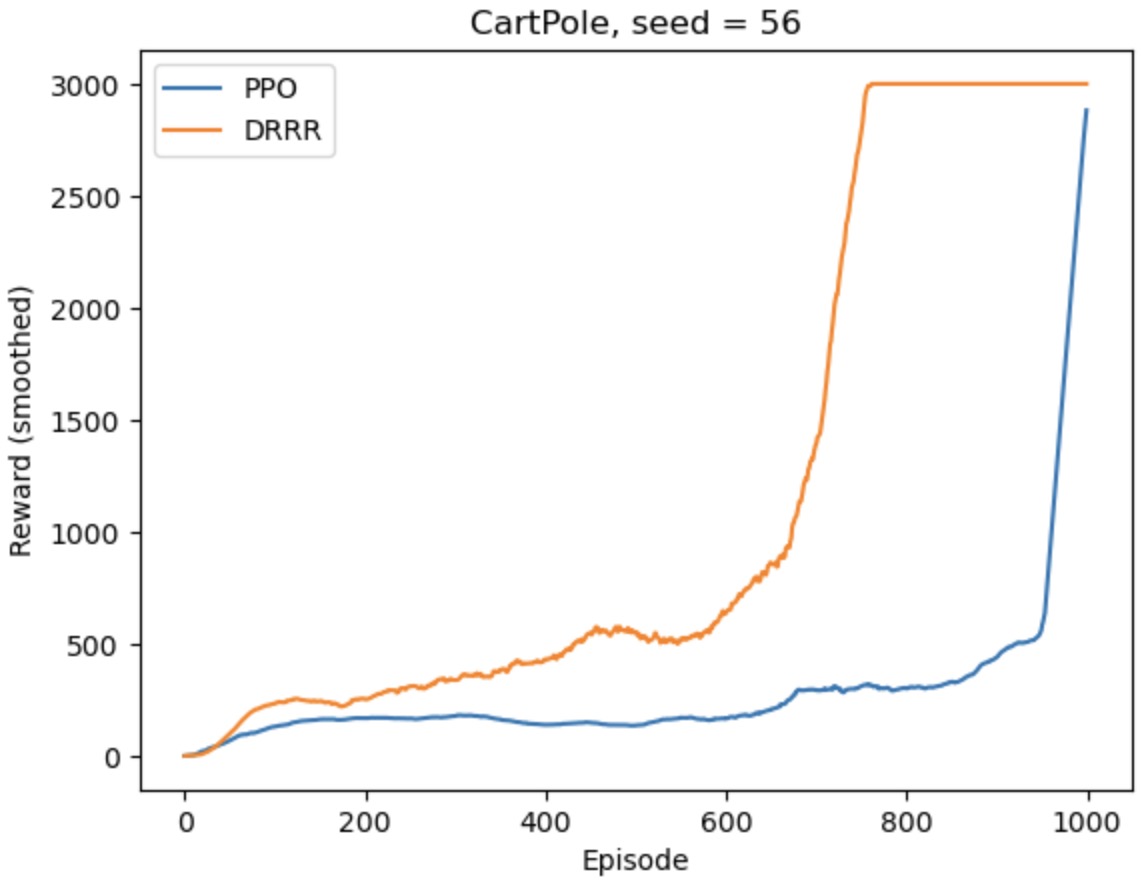}
        \caption{seed = 56}
        \label{fig:sub7}
    \end{subfigure}
    \hfill
    \begin{subfigure}{0.25\textwidth}
        \centering
        \includegraphics[width=\linewidth]{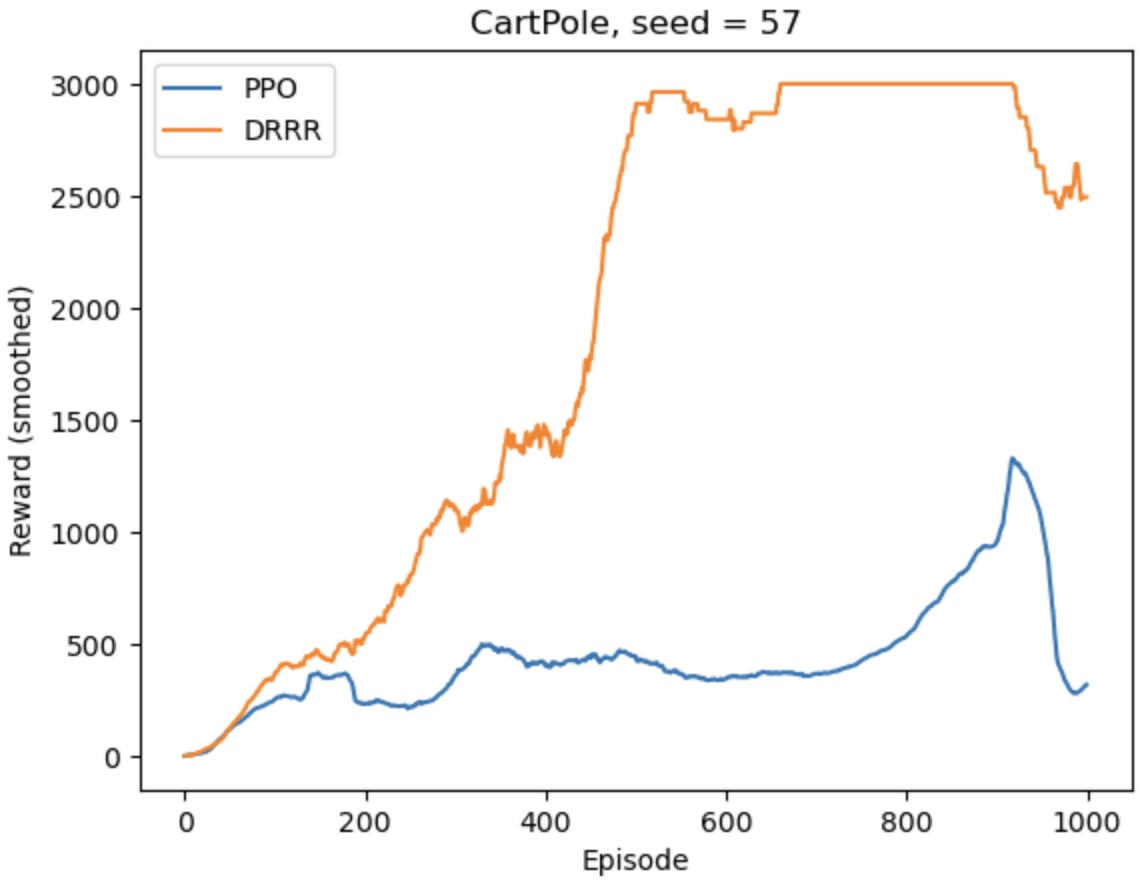}
        \caption{seed = 57}
        \label{fig:sub8}
    \end{subfigure}
    \hfill
    \begin{subfigure}{0.25\textwidth}
        \centering
        \includegraphics[width=\linewidth]{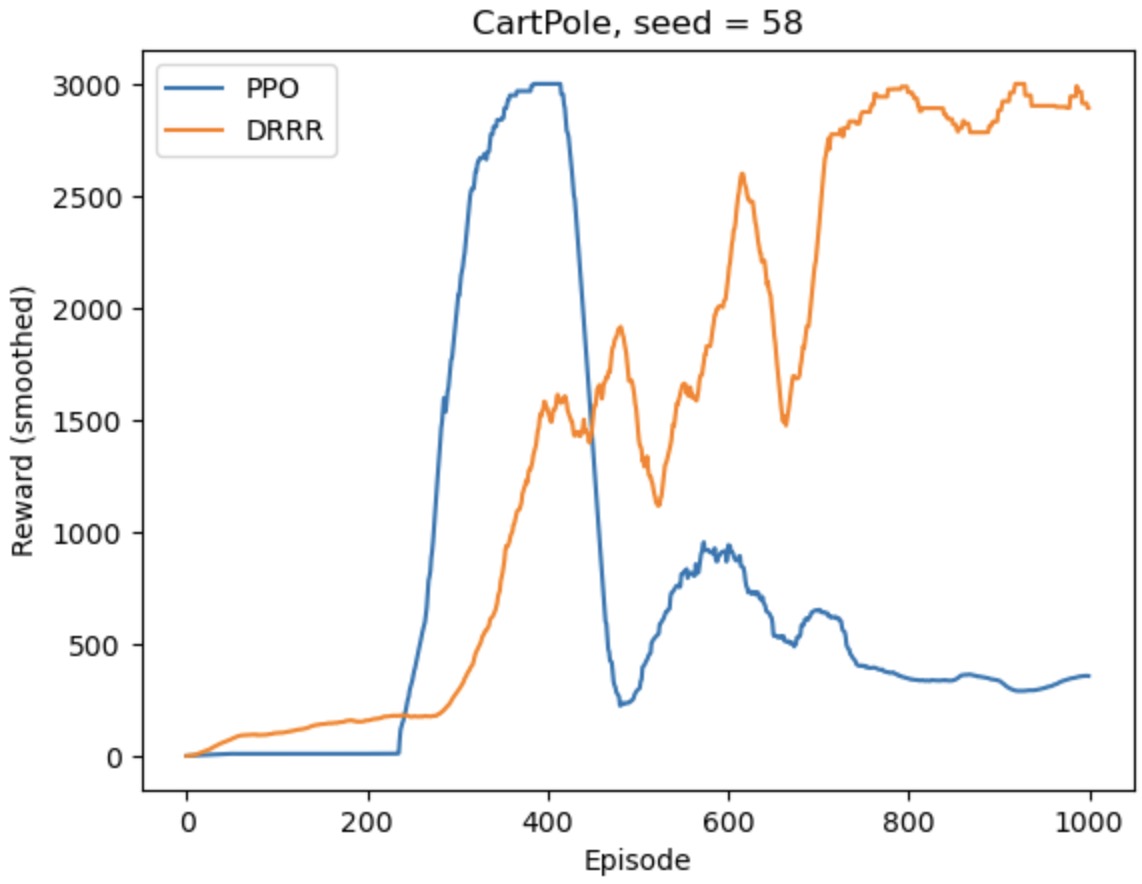}
        \caption{seed = 58}
        \label{fig:sub9}
    \end{subfigure}
    \caption{Performance of DR3 vs PPO agent on CartPole environment}
    \label{fig:grid5}
\end{figure}

\section{Future work}

There's a lot of additional work that can be done to improve R3 not implemented in the scope of this paper. For example, since the only responsibility of initiator is to find the first successful trajectory, the type of the initiator agent does not matter as much. As long as the agent can produce a probability for each action, we are safe to use that agent as an initiator. For example, we can experiment with replacing the initiator with a DDQN agent. We can also drop the initiator and make R3 a learning by demonstration algorithm using successful trajectory performed by human beings (and assign an uniform probability to each action), in which case we think the existence of replay buffer would significantly improve the performance.

There are also possible adjustments that can be done on the exploiter.
For example, there is still a chance that the exploiter can overfit to a certain trajectory in the replay buffer. The possibility of overfitting stems from if all of the probabilities in the trajectory are too high, then the trajectory will never be thrown out of the replay buffer due to low fitness.
As a result, we can limit the number of usage of each trajectory in the replay buffer to avoid this overfitting.

\section{Acknowledgement}

We would like to thank Prof. Pulkit Agrawal, our 6.8200 (Computational Sensorimotor Learning) instructor at MIT, for teaching us materials about reinforcement learning.
Also, we want to thank the 6.8200 course itself for providing us the code of PPO and DDQN algorithms.
Finally, we would like to thank Idan Shenfeld for providing us advice on how to improve our paper.



\appendix

\section{Appendix}

\subsection{Environment specs}

\begin{table}[!htbp]
\centering
\begin{tabular}{@{}ll@{}}
\toprule
\textbf{Component} & \textbf{Description} \\ \midrule
\textbf{Observation Space} & 
\begin{tabular}[c]{@{}l@{}}
- Current direction (up down left right) \\
- Current state of the grid (image of size * size * 3)
\end{tabular} \\ \midrule

\textbf{Action Space} & 
\begin{tabular}[c]{@{}l@{}}
Turn left, turn right, forward
\end{tabular} \\ \midrule

\textbf{Rewards} & 
\begin{tabular}[c]{@{}l@{}}
+1 for reaching the destination, 0 otherwise
\end{tabular} \\ \bottomrule
\end{tabular}
\caption{Details for the Crossing Environment}
\label{tab:minigrid-crossing-env}
\end{table}

\begin{table}[!htbp]
\centering
\begin{tabular}{@{}ll@{}}
\toprule
\textbf{Component} & \textbf{Description} \\ \midrule
\textbf{Observation Space} & 
\begin{tabular}[c]{@{}l@{}}
- Current direction (up down left right) \\
- Current state of the grid (image of size * size * 3)
\end{tabular} \\ \midrule

\textbf{Action Space} & 
\begin{tabular}[c]{@{}l@{}}
Turn left, turn right, forward, pick up an object, toggle an object
\end{tabular} \\ \midrule

\textbf{Rewards} & 
\begin{tabular}[c]{@{}l@{}}
+1 for reaching the destination, 0 otherwise
\end{tabular} \\ \bottomrule
\end{tabular}
\caption{Details for the Door Key Environment}
\label{tab:door-key-env}
\end{table}

\begin{table}[!htbp]
\centering
\begin{tabular}{@{}ll@{}}
\toprule
\textbf{Component} & \textbf{Description} \\ \midrule
\textbf{Observation Space} & 
\begin{tabular}[c]{@{}l@{}}
- Cart Position: The cart's position along the track\\ 
- Cart Velocity: The speed of the cart along the track\\ 
- Pole Angle: The angle of the pole with the vertical\\ 
- Pole Velocity At Tip: The rate of angle change at the tip of the pole
\end{tabular} \\ \midrule

\textbf{Action Space} & 
\begin{tabular}[c]{@{}l@{}}
Push cart to the left, push cart to the right
\end{tabular} \\ \midrule

\textbf{Rewards} & 
\begin{tabular}[c]{@{}l@{}}
+1 for every time step taken
\end{tabular} \\ \bottomrule
\end{tabular}
\caption{Details for the Cartpole Environment}
\label{tab:cartpole-env}
\end{table}

\subsection{Hyperparameters for R3 and DR3}

For both R3 and DR3, we are using importance sampling discard threshold $\sigma = 2$.
For R3, we are using bad fit discard threshold $\vartheta = 0.77 + 0.023 \cdot \len(\mathcal{B})$.
For DR3, we are using $\vartheta = 0.6$, and after each new trajectory has been added to the replay buffer, all old trajectories' threshold are increased by 0.02.
Also, after each time of usage, the threshold is increased by 0.01.

\end{document}